\documentclass[twoside,11pt]{article}
\PassOptionsToPackage{hyphens}{url}
\usepackage[hyphens]{url}
\usepackage{hyperref}
\hypersetup{
  breaklinks=true
}
\usepackage{jair, theapa, rawfonts}
\usepackage{graphicx}
\usepackage{amssymb}
\usepackage{subcaption}
\usepackage{amsthm}
\usepackage{amsbsy}
\usepackage{amsmath}
\usepackage{bbold}
\usepackage{multirow}
\usepackage{booktabs}
\usepackage{color}
\usepackage{float}
\usepackage{pifont}

\newsavebox{\tempfig}
\DeclareMathOperator*{\argmax}{argmax} 

\jairheading{80}{2024}{1033-1062}{09/2023}{07/2024}
\ShortHeadings{Does CLIP Know My Face?}
{Hintersdorf, Struppek, Brack, Friedrich, Schramowski \& Kersting}
\firstpageno{1033}

\begin{document}

\title{Does CLIP Know My Face?}

\author{%
    \name Dominik Hintersdorf \email Dominik.Hintersdorf@dfki.de \\
    \name Lukas Struppek \email Lukas.Struppek@dfki.de \\
    \name Manuel Brack \email Manuel.Brack@dfki.de \\
    \addr German Research Center for Artificial Intelligence (DFKI), \\
    Technical University of Darmstadt
    \AND
    \name Felix Friedrich \email friedrich@cs.tu-darmstadt.de \\
    \addr Technical University of Darmstadt, \\
    Hessian Center for AI (hessian.AI)
    \AND
    \name Patrick Schramowski \email Patrick.Schramowski@dfki.de \\
    \addr German Research Center for Artificial Intelligence (DFKI), \\
    Technical University of Darmstadt, \\
    Hessian Center for AI (hessian.AI), \\ 
    Ontocord
    \AND
    \name Kristian Kersting \email kersting@cs.tu-darmstadt.de \\
    \addr Technical University of Darmstadt, \\
    German Research Center for Artificial Intelligence (DFKI), \\
    Centre for Cognitive Science of Darmstadt, \\ 
    Hessian Center for AI (hessian.AI)
}

\maketitle

\begin{abstract}
\noindent
With the rise of deep learning in various applications, privacy concerns around the protection of training data have become a critical area of research. Whereas prior studies have focused on privacy risks in single-modal models, we introduce a novel method to assess privacy for multi-modal models, specifically vision-language models like CLIP. The proposed Identity Inference Attack (IDIA) reveals whether an individual was included in the training data by querying the model with images of the same person. Letting the model choose from a wide variety of possible text labels, the model reveals whether it recognizes the person and, therefore, was used for training.
Our large-scale experiments on CLIP demonstrate that individuals used for training can be identified with very high accuracy. We confirm that the model has learned to associate names with depicted individuals, implying the existence of sensitive information that can be extracted by adversaries. Our results highlight the need for stronger privacy protection in large-scale models and suggest that IDIAs can be used to prove the unauthorized use of data for training and to enforce privacy laws.
\end{abstract}

\section{Introduction}\label{sec:introduction}
\begin{figure*}
    \center
    \includegraphics[width=.80\textwidth]{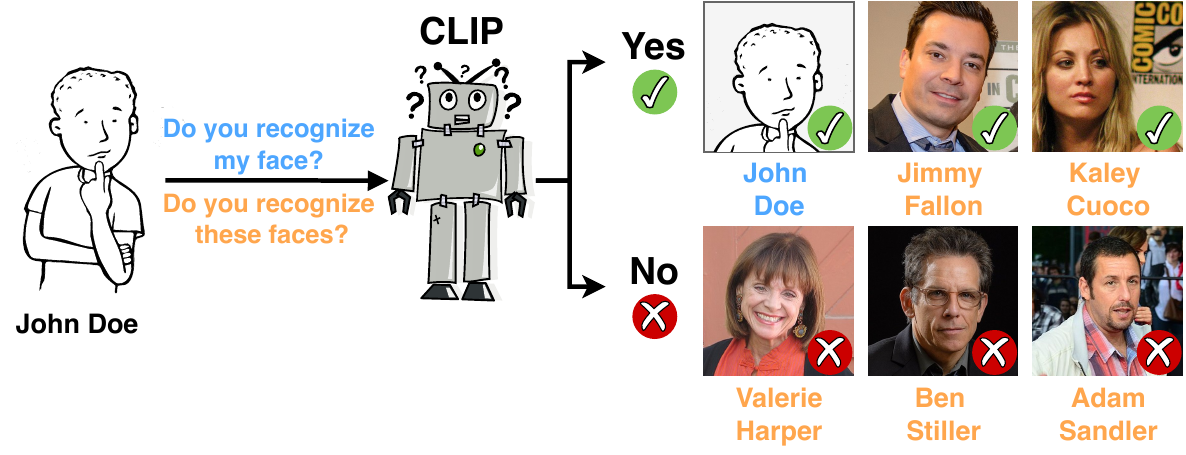}
    \caption{Illustration of our Identity Inference Attack (IDIA). True-positive (\ding{51}) and true-negative (\ding{55}) predictions of individuals to be part of the training data of CLIP. The IDIA was performed on a CLIP model trained on the Conceptual Captions 3M dataset~\cite{cc3m} where each person appeared only 75 times in a dataset with a total of 2.8 million image-text pairs. Images licensed as CC BY 2.0~\cite{jimmy_fallon,kaley_cuoco,ben_stiller,adam_sandler,valerie_harper}}
    \label{fig:teaser}
\end{figure*}

In recent years, models trained on massive datasets have shown impressive capabilities. 
Models like InstructGPT (ChatGPT)~\shortcite{instruct_gpt}, GPT-3~\shortcite{gpt3} and GPT-4~\shortcite{gpt4} not only caused a lot of excitement in the AI community but also got a lot of attention in the press and the broader society. As these models are trained on data scraped from the internet, people start to worry that those models memorize sensitive data during training. This is problematic in many ways. Not only can information like profession, place of residence, or even family status of individuals be extracted from models like GPT-3, as \citeA{what_does_gpt3_know} has shown in her MIT Technology Review Article “What does GPT-3 'know' about me?”, but these models can also make up false claims or fabricate facts about innocent people. As an example, the former Dutch politician Marietje Schaake was called a “terrorist” by Meta's BlenderBot~\cite{what_does_gpt3_know}.
While these large language models are trained only on text, research has started to focus more on the combination of multiple modalities, such as text and images, to create even more powerful models.
As a result, large multi-modal models are pushing current benchmarks in many applications like text-to-image synthesis~\shortcite{dalle,dalle2,glide,imagen,parti}, visual question answering~\shortcite{eichenberg2021,flamingo}, image captioning~\cite{clip_captioning} or even content moderation~\cite{schramowski2022}. 
Many of these applications rely on a text-encoder that was trained 
using \textbf{C}ontrastive \textbf{L}anguage-\textbf{I}mage \textbf{P}re-training~\shortcite{clip_radford}, or CLIP in short. Instead of requiring large labeled datasets to pre-train a model, CLIP uses image-text pairs collected from the internet to simultaneously train a vision and a text model in a contrastive learning setup. As a result, CLIP learns the visual concepts depicted in images through their accompanying texts and can be used in many downstream tasks, such as text-to-image models.
However, similar to text-only generative models, text-guided image generation models can also produce potentially defamatory images with faces of real private individuals.
Consequently, investigating which information is learned by these models and how it could be extracted is of large importance.

\begin{figure*}[ht]
    \centering
    \includegraphics[width=.60\textwidth]{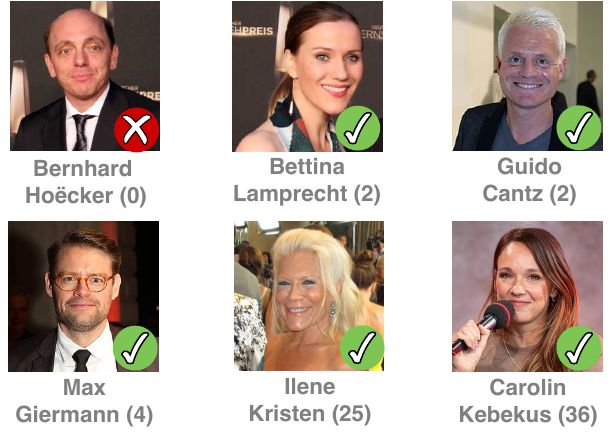}
    \caption{
        Identity Inference Attack Examples (IDIA) of infrequent appearing celebrities. True-positive (\ding{51}) and true-negative (\ding{55}) predictions of European and American celebrities in the LAION-400M dataset~\cite{laion400m}. The number in parentheses indicates how often the person appeared in the LAION dataset, containing 400 million image-text pairs. The IDIA was performed on a ViT-L/14 CLIP model trained on the LAION-400M dataset~\cite{open_clip}. Images licensed as CC BY 3.0~\cite{bernhard_hoecker,bettina_lamprecht,guido_cantz,max_giermann,carolin_kebekus} and CC BY 2.0~\cite{ilene_kristen}.
    }
    \label{fig:idia_example_laion}
\end{figure*}

These giant datasets used for training are usually collected under the assumption that any publicly available data scraped from the internet can be freely used without any restrictions.
However, data owners, including private individuals in particular, have never given consent for their data to be used to train these models. This is especially critical as these datasets often contain sensitive information including a person's name, address, or even sexual orientation and medical data~\cite{arstechnica_medical_imgs_laion}. Many legislators have recognized the potential threats to privacy and enacted laws to protect the personal data of their citizens, e.g., the General Data Protection Regulation (GDPR) in the European Union~\cite{gdpr_eu} or the Personal Information Protection and Electronic Documents Act (PIPEDA) in Canada~\cite{pipeda_canada}. The upcoming AI Act of the European Union is also subject to the GDPR, and with a few exceptions, AI systems will most likely also have to adhere to the GDPR. However, it remains unclear whether the current practice of using large datasets collected from the internet, which contain private information about individuals without them being aware, complies with existing privacy and copyright laws. Just recently, the stock image supplier \textit{Getty Images} started a lawsuit against \textit{Stability AI}, one of the creators of Stable Diffusion~\shortcite{rombach_diffusion}, over possible copyright infringement by scraping images from their site~\cite{getty_images_lawsuit_statement,getty_images_lawsuit_verge} and using them to train their generative text-to-image model.
In the past, there have already been comparable privacy incidents. For example, the company \textit{Clearview AI} was fined for scraping facial images for training their face recognition model~\cite{clearview_fine_uk,clearview_fine_italy,clearview_fine_greece}. Similarly, the companies \textit{Ever} and \textit{Weight Watchers/Kurbo} had to delete parts of their collected data, for training models on user data without the individuals' consent~\cite{ever_techcrunch,weight_watchers_ftc}. Current large-scale models keep pushing various benchmarks simply by scaling the dataset size. However, there is a major downside to this approach, as we demonstrate in this paper: larger datasets result in more information incorporated in the models, which in turn, leads to privacy breaches if private information has been part of the training data.

This highlights the necessity and urgency of reliable methods to trace the sources of training data and asses whether included individuals gave consent to use their data. Currently, companies seem to ignore these issues and individuals have to enforce their rights themselves, claiming their “right to be forgotten” (according to the GDPR~\cite{gdpr_eu}) and ensuring that their data is not used for future training.
A potent method to verify that specific data was used for training a machine learning model are membership inference attacks (MIAs). 
The goal of a MIA is to determine whether or not a particular data point, such as an image, was included in a model's training data. 
MIAs, however, are not only a tool but also pose a serious privacy risk.
If, for example, a model is trained on the medical data of patients with a specific disease, being part of the data can already pose a serious privacy leakage as this reveals a person suffering from this disease.
Being included in the training data of vision-language models like CLIP and the model having learned the faces and names of individuals and possibly other sensitive information poses additional severe privacy risks. For example, since CLIP is used in downstream tasks like generative text-to-image models, defamatory visual material of individuals could be created by just including their names in the prompts.

We propose a novel class of attacks for vision-language models, called \textbf{Identity Inference Attack (IDIA)\footnote{pronounced \textit{idea}}} to assess the information leakage and, therefore, the privacy risk of vision-language models. In this work, we investigate the question: “Does CLIP know my face?” or even the face of other individuals? 
Previous research on MIAs has focused primarily on uni-modal models and vision-language models like CLIP have not been investigated thoroughly in the context of MIAs and resulting privacy issues. 
Contrary to MIAs that are used to infer whether a specific sample was used for training a model, our IDIA takes an even broader view and asks whether data of a particular person was part of the training data. 
While such an attack poses a privacy risk with an attacker being able to deduce information about an individual, we believe that IDIAs can be used by individuals to detect whether they are part of private datasets of companies and thereby request their data to be deleted.

To summarize, this paper focuses on the memorization of vision-language models and investigates whether these models leak information about their training data. Specifically, we make the following contributions:

\begin{enumerate}
    \item We propose a new kind of privacy attack on vision-language models, called Identity Inference Attack (IDIA), which tests whether data of a person was used for training.
    \item Exhaustive experiments on large-scale datasets demonstrate that an attacker can infer with high accuracy whether a person was used for training, and that vision-language models indeed leak information about their training data.
    \item The IDIA shows that vision-language models are so powerful that they easily learn to connect a name with its corresponding face of depicted individuals during training, even if the person appears only a few times in the training data. This may imply a novel trade-off: the more powerful and “emergent” the abilities of a model, the greater the privacy leakage.  
\end{enumerate}

We proceed as follows. We start off by touching upon related work and background in Section~\ref{sec:related_work}. In Section~\ref{sec:identity_inference_attack}, we introduce our Identity Inference Attacks, followed by our experimental setup in Section~\ref{sec:experimental_setup}. After presenting our exhaustive empirical evaluation in Section~\ref{sec:experimental_results}, we discuss our results and possible implications in Section~\ref{sec:discussion}, before concluding in Section~\ref{sec:conclusion}. 

\section{Background and Related Work}\label{sec:related_work}
To enter the topic of privacy in AI, we start by giving an overview of CLIP. We then introduce related attacks on machine learning systems and discuss current research on the privacy of multi-modal systems.

\subsection{Vision-Language Models as Zero-Shot Classifiers}
Multi-modal learning describes the process of training machine learning models to process information from different modalities, e.g., written text, images, or audio.
Instead of training a standard image classifier directly on a labeled training set, the approach of \citeA{clip_radford}, called CLIP, introduced a novel contrastive pre-training scheme. 
For training, CLIP relies on public data scraped from millions of websites. 
More specifically, the training data consists of image-text pairs, where the images are accompanied by their corresponding captions. 
CLIP is trained by jointly optimizing an image encoder $M_{img}$ and a text encoder $M_{text}$, learning to match representations of images and the corresponding textual descriptions. This is achieved by forcing the calculated image and text embeddings to be close to each other for images with their correct caption while maximizing the distance to the remaining text embeddings in the current training batch.
While the vision model can be any standard neural network for computer vision tasks, for example, a ResNet~\shortcite{he2016} or a vision transformer (ViT)~\shortcite{vit}, a text-transformer~\shortcite{attention_is_all_you_need} architecture is usually used for the text encoder.
The original OpenAI CLIP model was trained on a non-public dataset with about 400 million image-text pairs. 
After training the multi-modal model on these image-text pairs, the authors demonstrated that CLIP achieves high zero-shot prediction performance on various tasks. 
Zero-shot prediction in this context means using the model for predictions on unseen datasets and tasks without fine-tuning it. Given a single image and multiple text prompts, the model will predict the text prompt that matches the visual concept of the image best.

For a prediction, the model is queried with an image $x$ and multiple text prompts $Z=[z_1, z_2, ..., z_n]$, containing descriptions of possible visual concepts, e.g., “an image of a horse” or “an image of a dog”. 
Based on the computed embeddings of both input modalities, the pairwise cosine similarity score between the embeddings of every text prompt~$z_i$ and the input image~$x$ is computed and scaled by a temperature term $\tau$:
\begin{equation}
    sim(x, z_i) = \frac{M_{img}(x) * M_{text}(z_i)}{\|M_{img}(x)\|_2 * \|M_{text}(z_i)\|_2} * e^{\tau}.
\end{equation}
Finally, the cosine similarities are normalized, converted into a probability distribution by a softmax function, and the prompt with the highest probability is predicted to be the best-matching text representation for the image.

This makes CLIP a versatile zero-shot classifier since the text labels and the number of potential classes for the classification can be chosen arbitrarily by defining additional concepts and classes with simple textual descriptions. 
Radford et al.~\cite{clip_radford} further demonstrated that carefully selected text prompts, also known as \textit{prompt engineering}, often improve the zero-shot prediction accuracy. 
For example, instead of using “\texttt{cat}” as a text prompt, embedding it into the prompt “\texttt{a photo of a cat}” improves the contextual understanding of the model. Even though these results are impressive, large-scale pre-training has, as we will show in this paper, also its downsides in terms of privacy.

\subsection{Privacy of Machine Learning Models}
Over time, various privacy attacks against machine learning systems have been proposed, which are oftentimes closely related to security attacks. 
Best known in the security field are adversarial attacks~\shortcite{szegedy_2014,adv_attacks_goodfellow}, which try to fool a model with hidden noise patterns not visible to the human eye. This pattern leads to the model predicting a potentially incorrect class. As recent work has shown, these attacks are not only of a theoretical nature but also affect production-ready systems~\shortcite{struppek_neuralhash,milliere2022}.
Related to adversarial attacks but more focused on the privacy aspect of machine learning models are model inversion attacks~\cite{fredrikson2015}. Their goal is to reconstruct specific training samples~\shortcite{zhang2020} or class representatives~\shortcite{wang2021,struppekModelInv} based on a model's learned knowledge.

Closer related to IDIAs for vision-language models are membership inference attacks (MIAs)~\shortcite{salem2019,yeom2018,carliniMIA,choquette2021,li2021,hintersdorfMIA} on uni-modal models, first introduced by \shortciteA{shokri2017}. 
By performing MIAs, the adversary aims to identify those samples from a data pool that were used for training a machine learning model. Successfully inferring this information indeed leads to a privacy breach, especially in domains like medicine or facial recognition. 
Whereas the literature on MIAs attests high success rates for inferring the membership status of specific samples, \citeA{hintersdorfMIA}, and \citeA{rezaei2021difficulty} have criticized that these attacks often have a high false-positive rate and thus their significance remains limited.

Moving beyond MIAs but staying uni-modal, \citeA{li2022userlevel} observed that embeddings of a person used for training an image classifier result in more compact clusters than images of other individuals in metric embedding learning.
Likewise, other approaches on text~\cite{song_user_level_text} and audio data~\shortcite{maio_user_level_audio} exploited the uniqueness of the data samples of a person to check whether their data was used for training.
For generative adversarial networks (GANs) \shortciteA{this_person_exists} have shown that even though generated images are believed to be novel, these images are very similar to training images and that an attacker can infer whether a person was used to train a GAN.
Finally, investigating the privacy of vision models trained with contrastive learning, \shortciteA{encodermi} have shown that these models are equally susceptible to MIAs. Their attack exploits the fact that during training, the augmentations of single images are close to each other in the embedding space, while the augmented images of samples not used for training are more distant from each other. 
Even though they looked at vision models pre-trained using CLIP, they only reviewed the privacy of the learned vision model by using classical MIAs, limited to specific data points, and did not investigate whether data of a person, in general, was used. While these works investigated privacy in a uni-modal setting, with the increasing popularity of multi-modal models, the demand for multi-modal privacy emerges at the same time.

\subsection{Privacy of Multi-Modal Systems}
The idea of combining different modalities to perform privacy attacks is not new. \citeA{multi_modal_emotion}, for example, have shown that multi-modal representations unintentionally encode additional information like gender or identity when recognizing emotions in speech. However, removing gender and speaker information from representations using adversarial learning seems to reduce information leakage. Similarly, \shortciteA{everything_about_you} have shown that by combining different modalities, like image and text information, locations, and hashtags in social media posts, an attacker can achieve higher accuracy in inferring relationships in social networks than when just using a single modality.
As multi-modal systems can also be used as data retrieval models to, e.g., search for images based on a text, \shortciteA{retrieval_pip} have proposed a privacy protection method to prevent an adversary searching for and combining private information in publicly released data such as photos. While training a retrieval system, the authors propose to modify the original data using adversarial perturbations to prevent images from being retrieved by an unauthorized party that does not have access to the trained system.

Recent works~\shortcite{meminf_on_tti,meminf_diffusion_models} on the privacy of diffusion models have shown that those models are, to a certain degree, also vulnerable to membership inference attacks. \shortciteA{carlini_diffusion} have even shown that isolated training images can be re-generated by large diffusion models, i.e., Stable Diffusion~\cite{rombach_diffusion} and Imagen~\cite{imagen}. 
As text-to-image models often rely on pre-trained multi-modal models like CLIP, privacy leakage could possibly also be attributed to CLIP.

To analyze how CLIP learns to abstract the visual concepts during training, \shortciteA{goh2021multimodal} have examined how neurons of CLIP's vision model react to certain stimuli. They have discovered that the model has neurons that are activated by different images that describe the same visual concept. As an example, the “Jesus Christ Neuron” is activated by pictures of Jesus Christ but also by pictures of other Christian symbols like crosses or crowns of thorns. This suggests that it is indeed possible that the model learns to associate different people with not only their names but also other attributes, potentially leading to privacy issues.
As a first proof-of-concept for this, \shortciteA{clip_me_if_you_can} demonstrated with a very limited example in a short blog post that CLIP does not only store general concepts of images and texts, but also some information about people's identities in the training data.
Similarly, the original CLIP paper~\cite{clip_radford} also noted that the model could be used to classify celebrities from the CelebA dataset~\cite{celeba}. However, to date, no systematically in-depth research has been conducted regarding the privacy of vision-language zero-shot classifiers such as CLIP. With our work, we aim to close this research gap and demonstrate that new privacy concerns go hand in hand with the rise of large-scale vision-language models trained on large web-crawled datasets. 
\newpage

\begin{figure*}[ht]
    \centering
    \includegraphics[width=\textwidth]{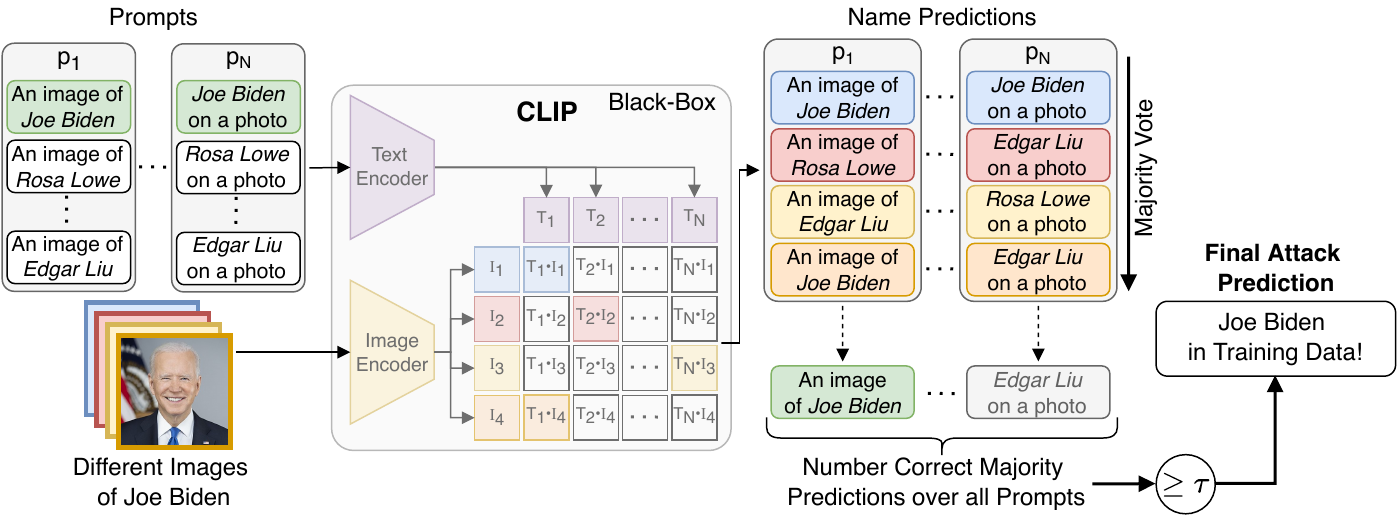}
    \caption{Identity Inference Attack (IDIA). Depiction of the workflow of our IDIA. Given different images and the name of a person, CLIP is queried with the images and multiple prompt templates containing possible names. After receiving the results of the queries, for each of the prompts, the name inferred for the majority of images is taken as the predicted name. If the number of correct predictions over all the prompts is greater or equal $\tau$, the person is assumed to be in the training data. (Best Viewed in Color)}
    \label{fig:entity_attack}
\end{figure*}
\section{Identity Inference Attack}\label{sec:identity_inference_attack}
After having provided sufficient background, we now introduce our Identity Inference Attack (IDIA). We provide an overview of the attack procedure in Fig.~\ref{fig:entity_attack}. In the following, we will refer to the person performing the IDIA as “adversary” in a game theoretical sense. In reality, the adversary can also be a person trying to assess whether he or she was in the training data.

\subsection{The Adversary's Capabilities}
As privacy attacks differ greatly with the capabilities and prior knowledge of the adversary, it is essential to outline the assumptions made for such an attack.
In our IDIA, we consider a user or adversary with closed-box access to a trained vision-language model (e.g., through an API), meaning that no access to the model's internals is given, preventing them from calculating any losses or other metrics. In literature, this is also called ``black-box'' access. However, to use more inclusive language, in this paper, we want to adhere to the IEEE Editorial Style Manual~\cite{ieee_style_manual}, which is why we will use the term closed-box. Performing privacy attacks in a closed-box setting is usually the hardest and most challenging scenario for the adversary. The number of queries the attacker can make to the model is not limited.
In our setting, the only prior knowledge of the adversary is possession of a certain number of images and the name of a person. Given that most people frequently upload images to social media together with their names, we believe that these assumptions are quite realistic. The adversary's goal is then to infer whether there are images in the model's training data that depict this particular person with a corresponding name description.\\

\subsection{Formalizing The Attack}
More formally, the adversary has access to a CLIP model $M_\mathit{CLIP}$ consisting of an image encoder $M_\mathit{image}$ and a text encoder $M_\mathit{text}$, embedding images and text into a joint embedding space. Additionally, a set $X=\{x_1, \ldots, x_M\}$ of facial images together with the real name $z_\mathit{real}$ of the person to perform the IDIA is available. 
The adversary further defines a large set of candidate names $Z=\{z_1,\ldots,z_K\}$ with $z_\mathit{real}\in Z$. 
These candidate names might contain the most common names that are culturally similar to $z_\mathit{real}$ and could, for example, be collected from public databases. Then, instead of querying the target model with plain names, a set of prompt templates $P=\{p_1,\ldots,p_N\}$ is defined, with each template being the style of “\texttt{a photo of \textit{$<$NAME$>$}}”. As \citeA{clip_radford} mentioned in their paper, such prompt engineering increases the performance of a zero-shot classifier like CLIP significantly and is common practice in current research~\shortcite{RiT,iPET}. 
Performing the IDIA using multiple different prompts increases the chance that captions, very similar to the used prompts, were part of the training data. 
Even though this probably will increase the success of the attack, we want to emphasize here that it is not a necessity that the exact prompts that are present in the training data are used for performing the IDIA.
For each template $p\in P$, the adversary creates $K$ prompts by replacing the placeholder \textit{$<$NAME$>$} with the $K$ different name candidates $z\in Z$. 
Subsequently, the model is queried with each combination of facial images $M$ and filled templates.
More specifically, for each prompt template $p_n$ and facial image $x_m$, the model predicts a name $\hat{z}_{m,n}$ as follows:
\begin{align*}
    \hat{z}_{m,n}= \argmax\limits_{z_k \in Z} \, sim(M_\mathit{image}(x_m), M_\mathit{text}(p_n \oplus z_k)).
\end{align*}
Here, $\oplus$ denotes the operation of replacing the \textit{$<$NAME$>$} placeholder in a prompt template with a name candidate. 
For each facial image $x_m$ out of $M$ images and a fixed prompt template $p_n$, the adversary obtains a name prediction $\hat{z}_{m,n}$. 
For the $M$ images available, this leads to a tuple $(\hat{z}_{1,n},\ldots,\hat{z}_{M,n})$ of name predictions. Out of this tuple, the most frequently predicted name is taken as the final name prediction $z^*_n$ for the prompt template $p_n$. 
Using the indicator function $\mathbb{1}$, which outputs $1$ if the correct name was predicted for a prompt template for the majority of input images and $0$ otherwise, the IDIA score $S$ is calculated as the sum of correct predictions over all $N$ prompt templates as follows:

\begin{align}
    \begin{split}
    S = \sum_{i=1}^N &\mathbb{1}(z^*_i,z_\mathit{real}), \\
    \text{with} \,\,  \mathbb{1}(z^*_i, z_\mathit{real})&:=\begin{cases}
        1, \quad \text{if} \,\, z^*_i=z_\text{real}\\
        0, \quad \text{otherwise}.
        \end{cases}
    \end{split}
\end{align}

To make a final prediction of whether the person was in the training set, the attacker applies a threshold function~$Q$ to the attack score $S$, which outputs $1$ if the real name has been predicted correctly for at least $\tau$ prompt templates, with $\tau \in \{1,\ldots,N\}$:

\begin{align}
    Q(S, \tau) =
    \begin{cases}
    1, &\text{if}\;S \geq \tau\\
    0, &\text{otherwise}.
    \end{cases}
\end{align}\\

In our experiments, we set the threshold to $\tau = 1$, meaning that the name has only to be correctly predicted for a single prompt template to conclude that the training data contained images of this person. As will be shown in our experimental results, this is already enough to achieve a very low false-positive rate. To reduce the risk of false-positive predictions by chance even more, one can define a large set of candidate names, increase $\tau$, and use more images for the attack.
At the same time, adding more prompts will most likely further increase the robustness and stability, and the attack will produce fewer false-negative predictions.

\section{Experimental Setup}\label{sec:experimental_setup}

We will now describe the experimental setup used to investigate the information leakage of CLIP with IDIA in more detail. This includes the analysis of the datasets for the experiments and the training of target models. Our code, as well as the pre-trained models, are publicly available on GitHub\footnote{\url{https://github.com/D0miH/does-clip-know-my-face}} and a demo of our IDIA can be found on Hugging Face\footnote{\url{https://huggingface.co/spaces/AIML-TUDA/does-clip-know-my-face}}. \\

To evaluate the information leakage of CLIP and the effectiveness of our attack, we consider CLIP models that were pre-trained on the LAION-400M~\cite{laion400m,open_clip} and the Conceptual Captions 3M~\cite{cc3m} (CC3M) datasets, containing image-text pairs. While the LAION-400M dataset contains 400 million, the CC3M dataset contains 2.8 million image-text pairs. For the identities, for which we want to infer whether they are part of the training data, we use images of male and female actors from the FaceScrub dataset~\cite{facescrub}, consisting of images of 530 identities, for several reasons. The first reason for choosing celebrities is that we need enough images of each person to evaluate the success of an IDIA thoroughly. The second reason is that we want to respect the privacy of private individuals, not being celebrities or of public interest. Using private individuals for these experiments could possibly violate their privacy. As will be shown later, it is very hard to find individuals of public interest that are not present in the LAION-400M dataset. Therefore, we perform more controlled experiments on the CC3M dataset, for which we can control the number of facial images together with a corresponding name description in the training data.

To correctly calculate evaluation metrics, including the true-positive and false-positive predictions, we first have to analyze which individuals are already part of the datasets and which are not. For the LAION-400M dataset, we search in all 400 million captions for the names of the individuals. 
Theoretically, there could exist individuals with the same name, confounding the analysis. However, as we use celebrities of whom it is more likely to find images online, we assume that the names of these individuals are distinct. 
Having analyzed the captions, we can tell how frequent each individual appears in the dataset. 
Because the adversary needs different images of the same person for the attack, we use only images of individuals in the FaceScrub dataset which have at least 30 images available, resulting in 507 different celebrities.
The result of the analysis is that all 507 celebrities are part of the LAION-400M dataset. As the image-text pairs are scraped mostly from English websites, we manually searched for 30 images of 9 European male and female actors, respectively.
With the expectation that these people will not appear in English-language TV series or films, it is less likely that they will be in the dataset and found by caption matching.
However, during the analysis, we found only 8 of the 18 European celebrities not to be part of LAION-400M. 
Even though 10 of the European celebrities can be found in the dataset, the frequency of their appearance is much lower than those of others, resulting in more meaningful experimental results for individuals with only a few appearances.
To avoid distorting the experimental results by using individuals who occur very frequently in the dataset, we only use individuals who are contained less than 300 times in the 400 million images of the dataset.
As a result of the whole dataset analysis, we use 152 individuals in total for the experiments, 144 of which are included in the dataset, while 8 individuals are not.

Analyzing the CC3M dataset is a bit more complicated, as Sharma et al.~\cite{cc3m} have anonymized named entities in the image captions by replacing them with their hypernym, e.g., celebrity names with “\textit{actor}”. While this preprocessing would already suffice for our experiments since we would like to explore whether vision-language models connect the names of individuals with their pictures, we want to reduce possible side effects of images appearing in the dataset without the corresponding name of the person.
Filtering the dataset by searching for the hypernyms or the names in the captions is not possible, as it is very hard to specify the exact hypernym for each of the celebrities' names.
This is why we analyzed the 2.8 million images of the CC3M dataset using facial recognition to check whether the celebrities are depicted on the images of the CC3M dataset. 
More details on how the analysis was performed can be found in App.~\ref{app:cc3m_analysis}. 
We then added image-text pairs to the CC3M dataset in a controlled way.
For this, we searched for similar image-text pairs in the LAION-5B dataset~\shortcite{laion5b}, for each of the 530 celebrities in the FaceScrub dataset.
After confirming that the names of the celebrities are present in the captions of the found images, we added the image-text pairs to the CC3M dataset to train CLIP models. In total, we used 200 individuals for our experiments on the CC3M dataset, 100 of which were added to the dataset and 100 were not used to train the models. Each group consists of an equal number of images depicting male and female actors.

Our experiments evaluate the IDIA on several CLIP models with different image feature extractors. We attacked the three CLIP models ViT-B/32, ViT-B/16, and ViT-L/14, all trained on the LAION-400M dataset by the OpenCLIP project~\cite{open_clip}. The models achieve an ImageNet top-1 zero-shot validation accuracy of 62.9\%, 67.1\%, and 72.77\%, respectively, which is comparable to the performance reported in Radford et al.~\cite{clip_radford}. This highlights that these models have learned meaningful visual concepts during training.
For models trained on the CC3M dataset, we used CLIP models with a ResNet-50~\cite{he2016}, ViT-B/32~\cite{vit}, and a ResNet-50x4~\cite{efficient_net} as image feature extractors.
During all experiments, the text encoder architecture of the CLIP models was identical to the one defined in the original CLIP paper~\cite{clip_radford}.
Because we want to investigate how the number of images of a person influences the attack success, we trained a total of 18 models on the CC3M dataset. For each of the ResNet-50, ViT-B/32, and ResNet-50x4 architectures, we trained the models with 75, 50, 25, 10, 5, and 1 data sample occurrences per person. On average, the ViT-B/32, ResNet-50, and ResNet-50x4 models achieve an ImageNet top-1 zero-shot validation accuracy of around 14\%, 19\%, and 21\%, respectively. 
The difference in validation accuracy between the models trained on LAION and these models can be attributed to the much lower number of data points in the CC3M dataset. Research by \shortciteA{mehdi_clip_scaling} has shown that scaling the model size and keeping the dataset size fixed still increases the zero-shot accuracy of CLIP models. This, and the fact that the validation accuracy of all models was consistently increasing throughout the training, shows that the models did not overfit on the CC3M dataset.
Details and hyperparameters about the training, as well as the number of parameters for all models, can be found in App.~\ref{app:target_models}.

For the IDIA, we crafted $21$ different prompt templates and used $1000$ possible names for the model to choose from. The prompt templates, as well as details on how the names were generated, can be found in App.~\ref{app:prompt_temp_name_gen}. As previously mentioned, we used up to 30 images of each person to perform the attack. We set the prompt threshold $\tau=1$ such that the person is predicted to be in the training data if the correct name is predicted for at least one prompt template. Keep in mind that the probability of predicting the correct name in this setting by chance for a given image is $\frac{1}{1000}$. Because the IDIA performs one prediction for each image for a given prompt, the probability of predicting the correct name for the majority of the given images is $(\frac{1}{1000})^k$ where $k$ is the number of images for which the name was predicted. This makes it highly unlikely that IDIA produces false-positive predictions when many images are used for the attack.

\section{Attack Results}\label{sec:experimental_results}

\begin{figure*}[ht]
    \centering
    \begin{subfigure}[b]{0.325\textwidth}
        \centering
        \includegraphics[width=\textwidth]{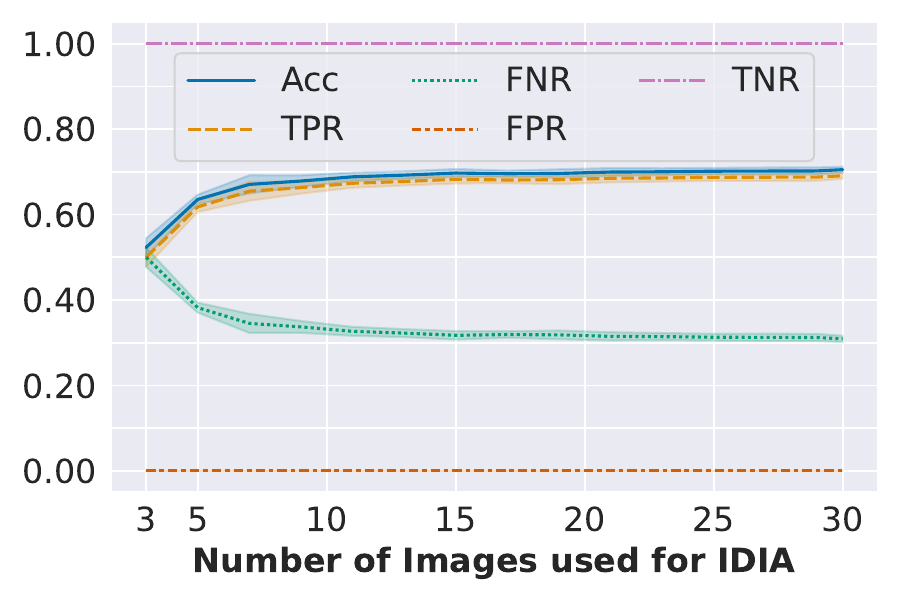}
        \caption{LAION-400M ViT-B/32}
        \label{fig:attack_sample_influence_laion_vitb32}
    \end{subfigure}
    \begin{subfigure}[b]{0.325\textwidth}
        \centering
        \includegraphics[width=\textwidth]{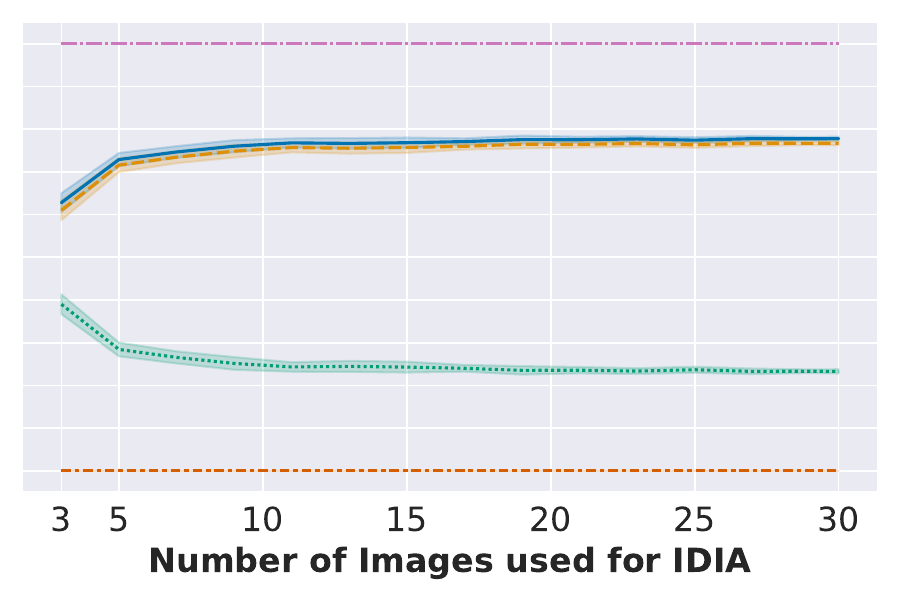}
        \caption{LAION-400M ViT-B/16}
        \label{fig:attack_sample_influence_laion_vitb16}
    \end{subfigure}
    \begin{subfigure}[b]{0.325\textwidth}
        \centering
        \includegraphics[width=\textwidth]{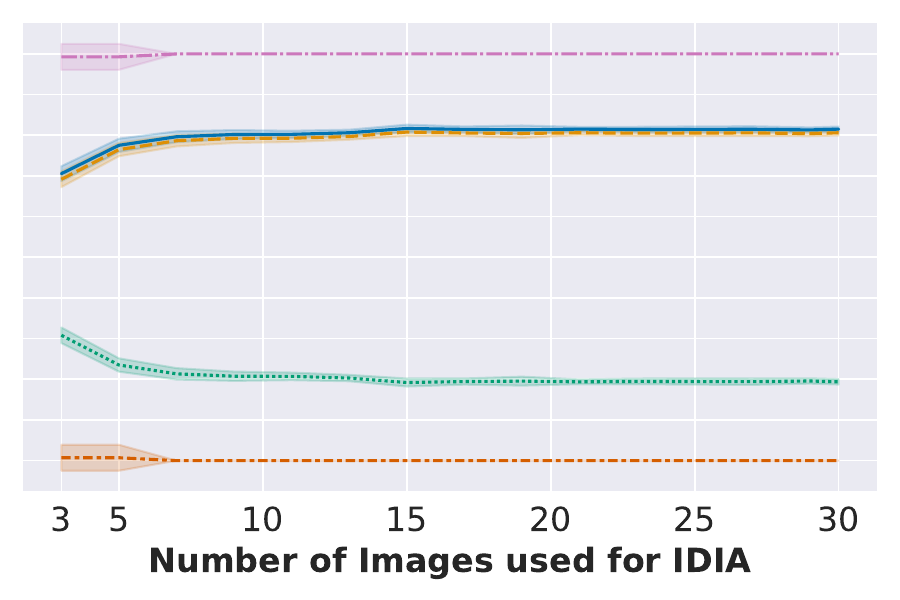}
        \caption{LAION-400M ViT-L/14}
        \label{fig:attack_sample_influence_laion_vitl14}
    \end{subfigure}
    \begin{subfigure}[b]{0.325\textwidth}
        \centering
        \includegraphics[width=\textwidth]{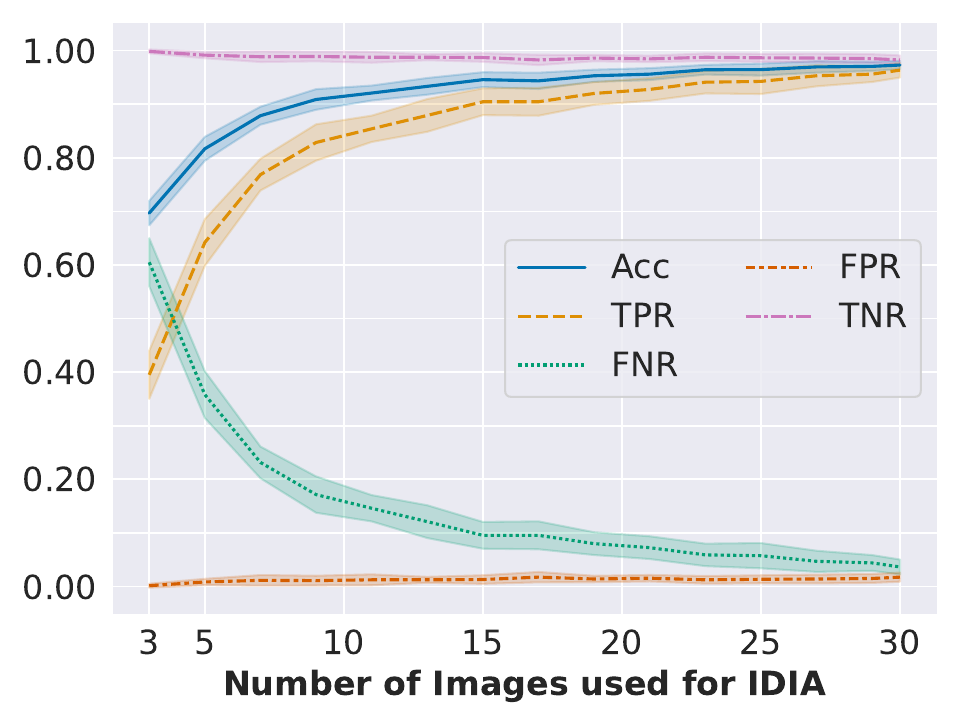}
        \caption{CC3M ResNet-50}
        \label{fig:attack_sample_influence_cc3m_rn50}
    \end{subfigure}
    \begin{subfigure}[b]{0.325\textwidth}
        \centering
        \includegraphics[width=\textwidth]{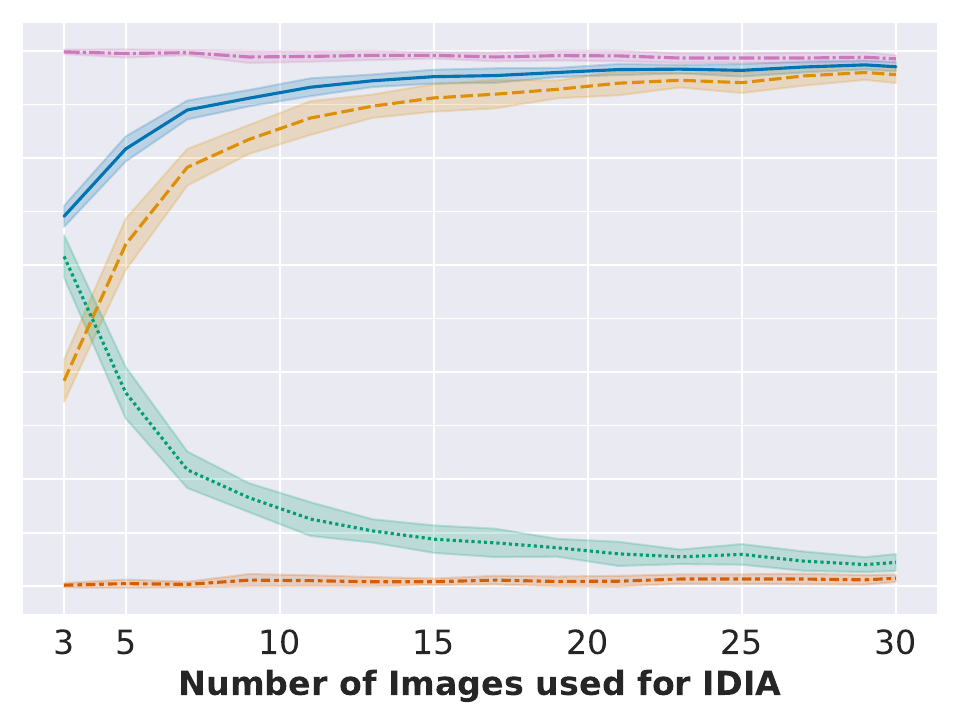}
        \caption{CC3M ResNet-50x4}
        \label{fig:attack_sample_influence_cc3m_rn50x4}
    \end{subfigure}
    \begin{subfigure}[b]{0.325\textwidth}
        \centering
        \includegraphics[width=\textwidth]{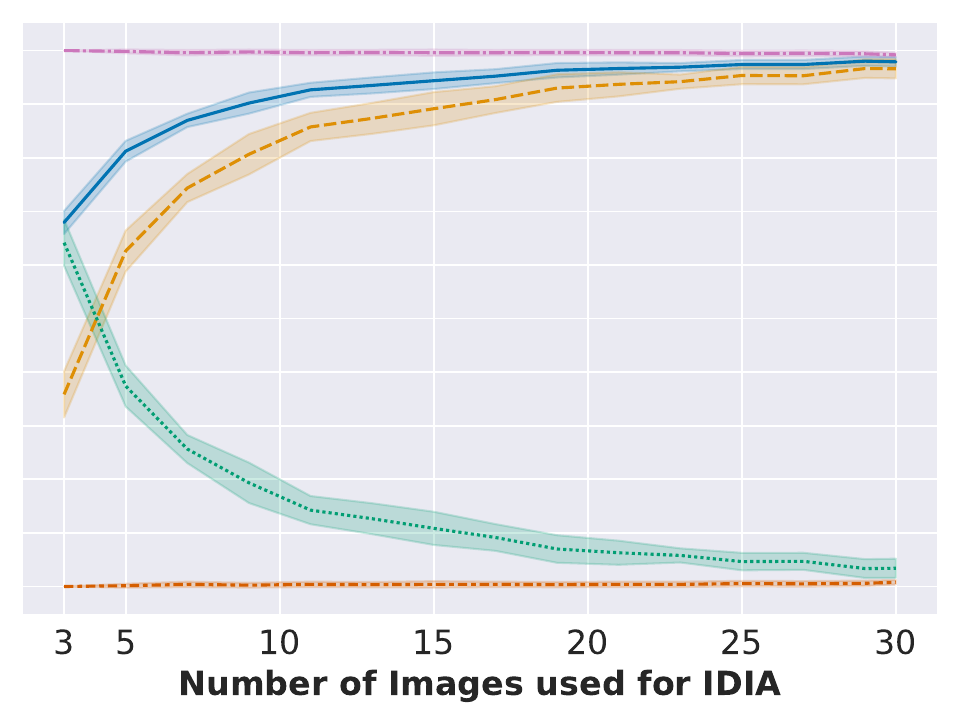}
        \caption{CC3M ViT-B/32}
        \label{fig:attack_sample_influence_cc3m_vitb32}
    \end{subfigure}
    \caption{IDIAs can already be executed with only a few samples. Depicted is the influence of the number of attack samples available to the adversary during the IDIA. The models were trained on the LAION-400M and the CC3M datasets. Plotted are the mean and standard deviation of the true-positive rate (TPR), false-negative rate (FNR), false-positive rate (FPR), true-negative rate (TNR), and accuracy (Acc). Metrics are computed by repeating the IDIA 20 times with a randomly sampled subset of attack samples. For the CC3M models, each individual was present 75 times in the dataset, while for the LAION-400M dataset each individual appeared less than 300 times.}
    \label{fig:attack_sample_influence}
\end{figure*}

After clarifying the experimental setup, we will now present the outcome of the experiments.
Examples of true-positive and true-negative predictions can be seen in Fig.~\ref{fig:teaser} and in Fig.~\ref{fig:idia_example_laion}.
The results for the IDIA with a varying number of attack samples available to the adversary and targeting different CLIP models can be seen in Fig.~\ref{fig:attack_sample_influence}. Because some individuals have more than 30 images available, the attack was repeated 20 times, with randomly sampling a subset of images per individual. In the following plots, the mean and standard deviation are reported. Detailed confusion matrices for each of the performed attacks with 30 attack samples can be found in App.~\ref{app:additional_experimental_results}.
As can be seen in Fig.~\ref{fig:attack_sample_influence}, for the models trained on LAION, the IDIA has an accuracy of more than 70\% when the individuals appeared only less than 300 times in the dataset containing 400 million data points, and 30 images are used for the attack. While the mean true-positive rate for the attacks on the LAION models is consistently above 69\%, using 30 images for the IDIA, the mean false-positive rate is 0\% in all cases.
Similarly, the true-positive rate for the IDIAs on the CC3M models is consistently above 95\% while the mean false-positive rate is below 1.8\%.
We have observed that our IDIA very rarely produces false-positives, increasing trustworthiness in the positive results of the attack.
For the models trained on CC3M it seems that the performance of the IDIA is logarithmically increasing with the number of samples used for the attack. As a result, using more than 20 samples looks like it has diminishing returns. In contrast to that, it appears that, in general, fewer samples are required for attacking the models trained on the LAION dataset than models trained on the CC3M dataset, even when attacking the same model architecture with the same number of trainable parameters (ViT-B/32).
Comparing the results of the experiments on the LAION models, one can clearly see that the success rate of the IDIA increases with model size. The ViT-L/14 model has more than triple the number of parameters for the vision model, while the number of parameters for the text transformer was more than doubled. 
As a result, the true-positive rate for the ViT-L/14 model is increased by 11.75\% and 3.6\% when compared to the ViT-B/32 and ViT-B/16 models, respectively.
Comparing the results of the different models trained on the CC3M dataset, it seems that on this dataset the model size has no impact on the performance of the attack. Even though the number of parameters for the vision model has more than halved and for the text encoder decreased by roughly 36\% from the ResNet-50x4 to the ResNet-50, there is no noticeable difference in the success of the attack evident. 
\begin{figure*}[ht]
    \centering
    \begin{subfigure}[b]{.40\textwidth}
        \centering
        \includegraphics[width=6.3cm]{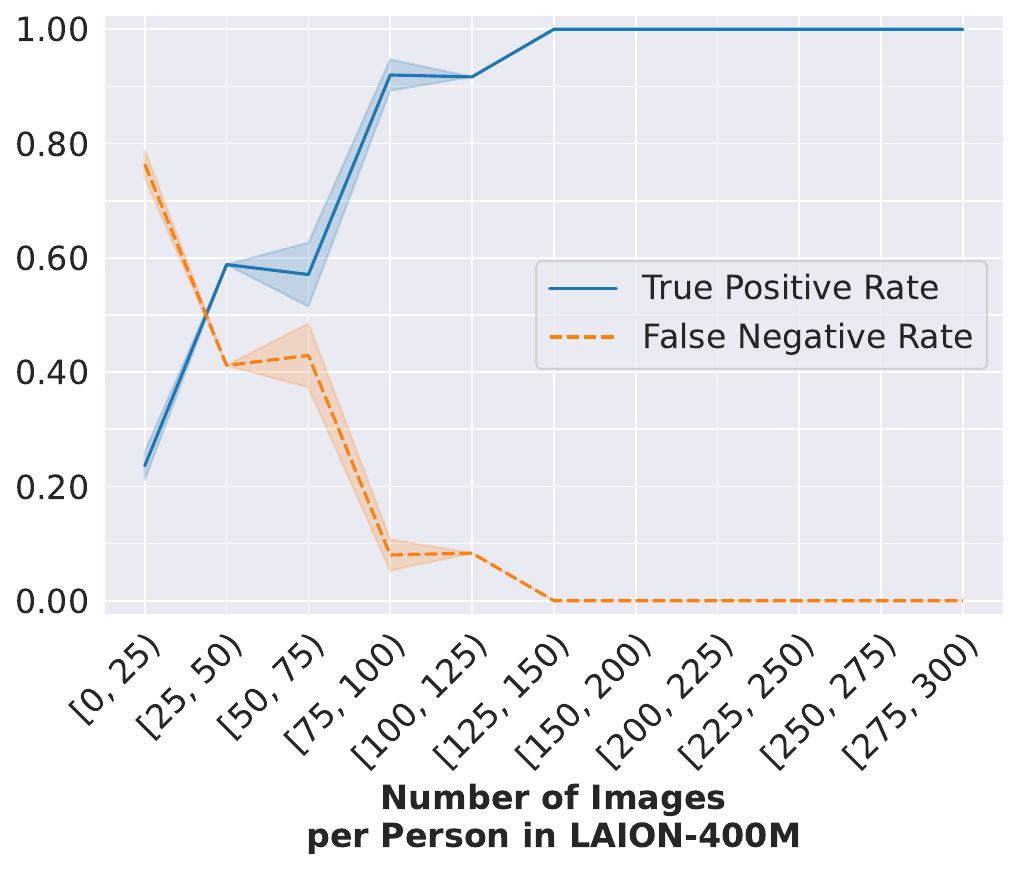}
        \caption{LAION-400M ViT-L/14}
        \label{fig:num_training_samples_laion_vitl14}
    \end{subfigure}
    \begin{subfigure}[b]{.35\textwidth}
        \savebox{\tempfig}{
            \includegraphics[width=6.3cm]{plots/num_training_samples/num_training_samples_plot_multiprompt_laion400_ViT-L-14_1000_300_1.pdf}
        }
        \raisebox{\dimexpr\ht\tempfig-\height}{
            \includegraphics[width=5.2cm]{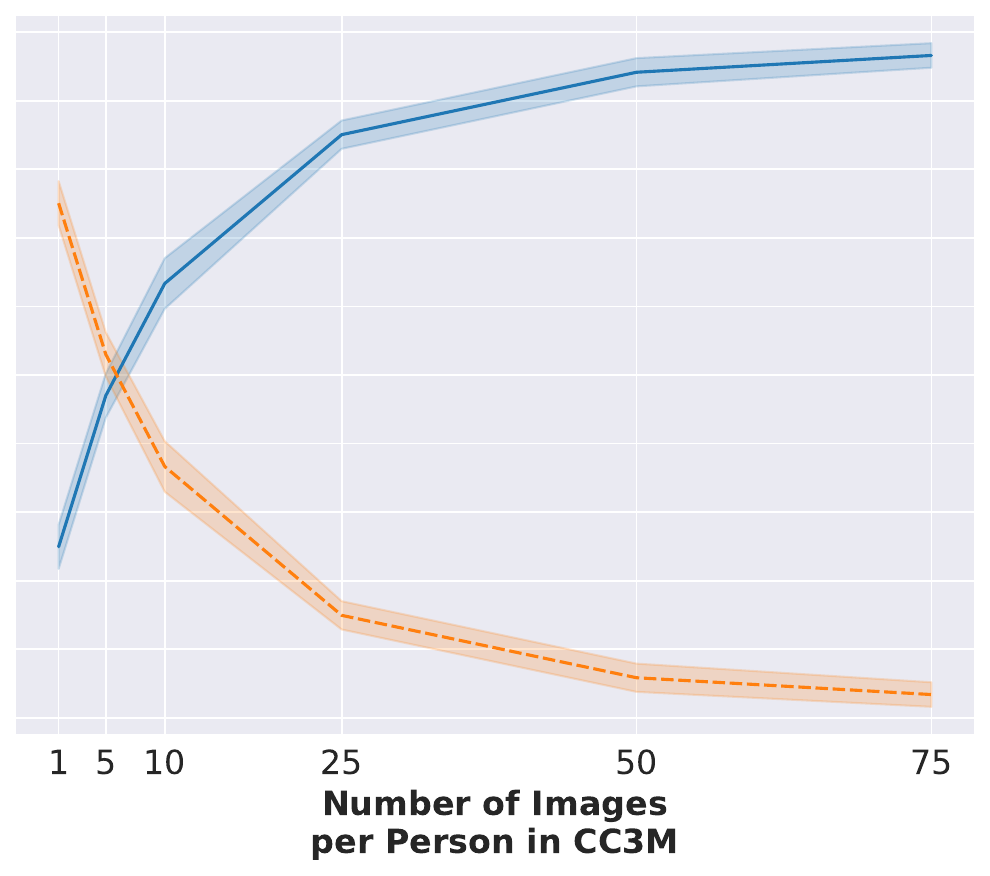}
        }
        \captionsetup{justification=centering}
        \caption{CC3M ViT-B/32}
        \label{fig:num_training_samples_cc2m_vitb32}
    \end{subfigure}
    \caption{IDIAs work even on individuals who appear very few times in the dataset. Depicted are the mean and standard deviation of the true-positive and false-negative rates of the IDIA for different numbers of training images per person. In this experiment, 30 samples were used to perform the attack. Additional plots for the other models can be found in App.~\ref{app:additional_experimental_results}}
    \label{fig:num_training_sample_influence}
\end{figure*}

In general, the lower the false-positive rate of a privacy attack, the more confident the adversary can be in the results. High false-positive rates, on the other hand, reduce the risks of privacy attacks since false-positive predictions could falsify the results. The very low false-positive rates in the experiments emphasize the fact that an adversary provided with only a few images of a person can already successfully and confidently infer whether someone's photos have been used to train a model. This indicates that the adversary can indeed infer sensitive information about the training data.
Interestingly, the specific sample selection does not seem to have a high impact on the results. While the standard deviation of the false-positive and true-negative rates is always below 0.85\%, the standard deviation of the false-negative and the true-positive rates is below 1.79\% in all experiments. This indicates that the selection of an individual's images used for the IDIA has almost no effect on the success of the attack.

To investigate how the number of images of a person present in the training data affects the success of the attack, we plot the true-positive and the false-negative rates for different numbers of training samples per person. The results of the ViT-L/14 trained on LAION and the different ViT-B/32 models trained on CC3M with 75, 50, 25, 10, 5, and 1 images per person present in the dataset are stated in Fig.~\ref{fig:num_training_sample_influence}.
The true-positive rate seems to increase logarithmically with the number of appearances of an individual in the training data. For the LAION dataset, we sort the people being present less than 300 times in the dataset into 12 bins according to their frequency of appearance.
As the individuals are not distributed evenly between the bins, the results vary more compared to the results of the models trained on CC3M, where the number of individuals is always the same.
For the LAION ViT-L/14 model, with only 25 to 50 images of a person among 400 million training samples, the IDIA can already correctly identify three out of five people whether their data was used to train a model.
On the CC3M dataset, with only 10 images of a person among 2.8 million training samples, an adversary can already correctly identify about two out of three people whether they were part of the training data. If the person appeared only a single time in the training data, the adversary can still predict the membership of the person's data with a true-positive rate of 24.9\% for the ViT-B/32 model trained on CC3M. 
The results for the other models trained on LAION and CC3M are very similar and can be found in App.~\ref{app:additional_experimental_results}.

\begin{figure*}[ht]
    \centering
    \begin{subfigure}[b]{0.49\textwidth}
         \centering
         \includegraphics[width=7cm]{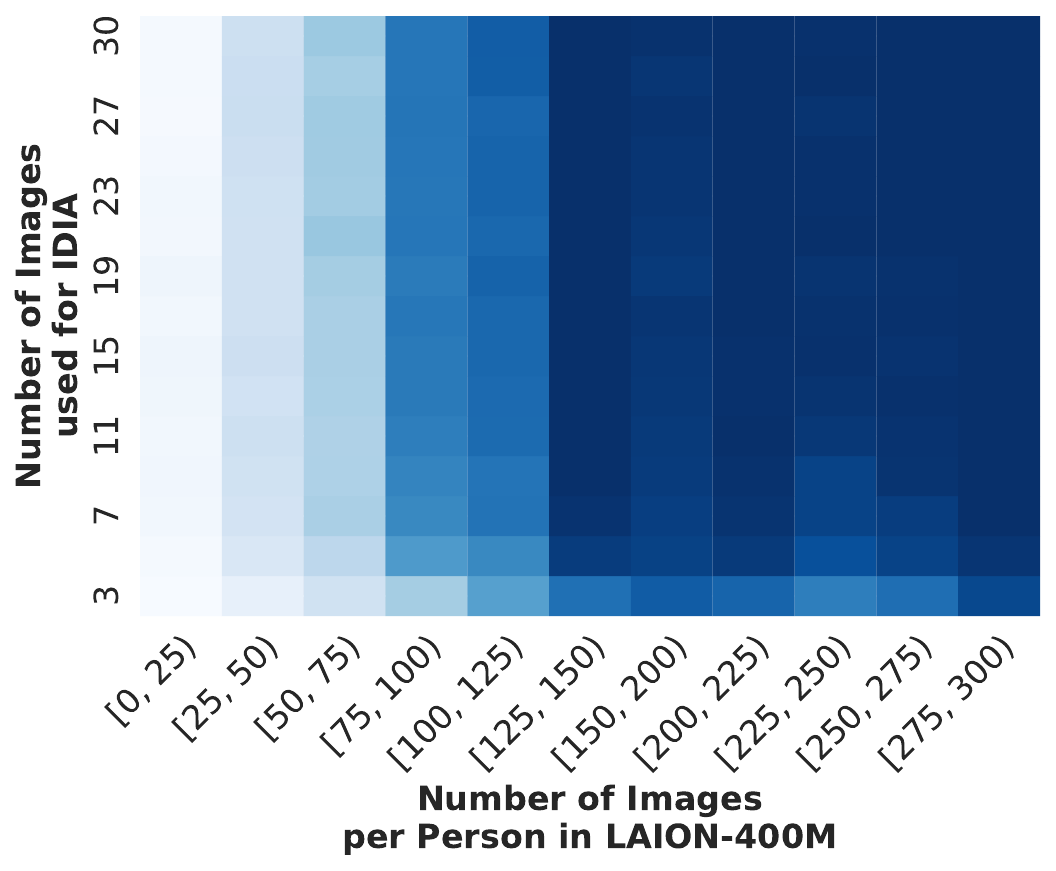}
         \caption{LAION-400M ViT-B/32}
         \label{fig:heatmap_cc2m_rn50}
     \end{subfigure}
     \begin{subfigure}[b]{0.49\textwidth}
         \centering
         \savebox{\tempfig}{
            \includegraphics[width=7cm]{plots/heatmaps/heatmap_num_training_samples_laion400_ViT-B-32_1000_300_1.pdf}
        }
        \raisebox{\dimexpr\ht\tempfig+0.1cm-\height}{
            \includegraphics[width=6.8cm]{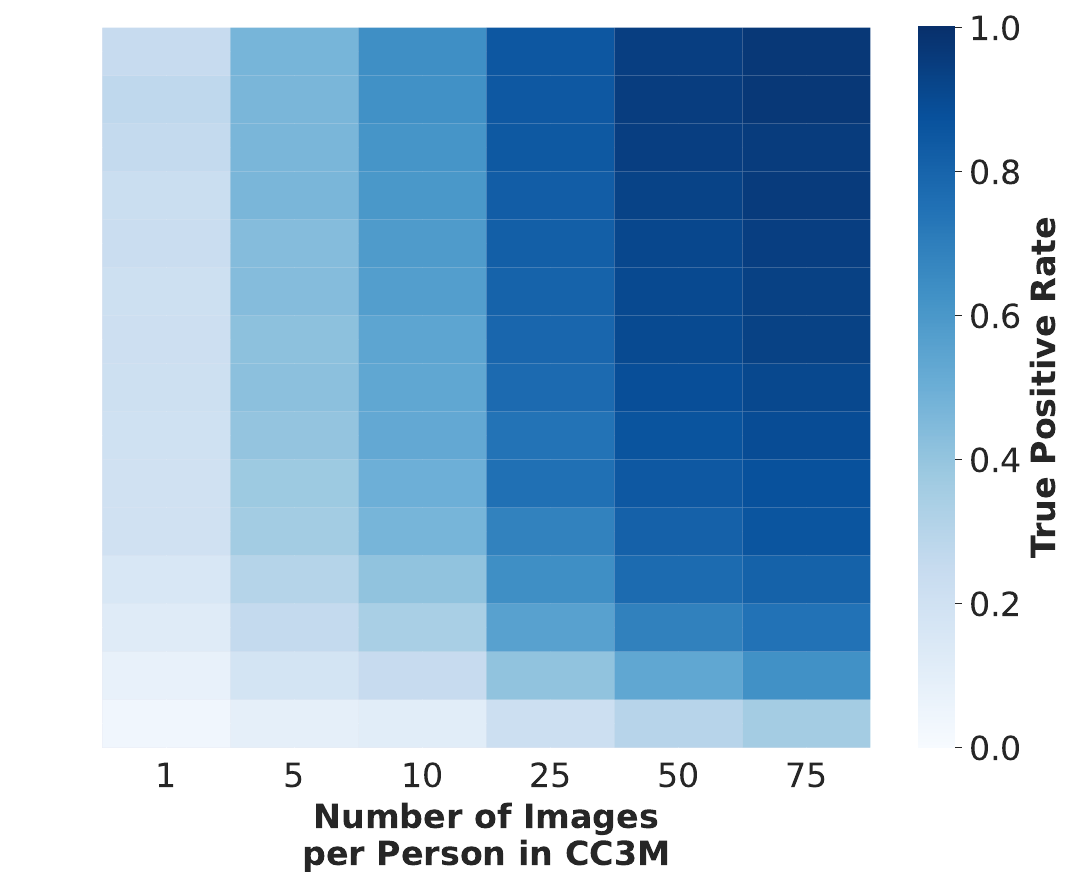}
        }
         \caption{CC3M ViT-B/32}
         \label{fig:heatmap_cc2m_rn50x4}
     \end{subfigure}
    \caption{The Influence of the number of samples available to the attacker and the number of samples per person in the training data on the true-positive rate of the attack. With a higher number of training samples per individual, the advantage of having more samples available for the attack is less pronounced for each of the models.}
    \label{fig:heatmap_cc2m}
\end{figure*}

As we have demonstrated, the number of samples used for the attack, as well as the number of samples used for training, influence the success of the IDIA. However, we hypothesize, that those two attack variables also influence each other. To answer the question of what influence the number of available samples has on the success of the attack, depending on the number of samples per person that were present in the training data, we plot the true-positive rate for varying combinations in Fig.~\ref{fig:heatmap_cc2m}.
As shown in the heatmaps, there appears to be a saturation point where the number of images used for the IDIA no longer seems to have a drastic effect. For the LAION ViT-B/32 model, this seems to be the case if more than 125 images are present in the training data, while for the CC3M ViT-B/32 model, this seems to be the case at more than 10 images.
Even though the true-positive rate increases with the number of samples, this effect is seemingly much less pronounced for a greater number of appearances in the training data.
However, comparing the LAION and CC3M experiments, the point of saturation appears to be very much dependent on the total size of the training data.

\section{Discussion}\label{sec:discussion}

\subsection{A Trade-Off between Power and Privacy}
In our experiments we have shown that models are memorizing sensitive parts of their training data, i.e., the names and the appearance of a person, which poses a grave threat to privacy.
Models like CLIP, for example, are used in many downstream tasks such as Stable Diffusion~\cite{rombach_diffusion} or DALL-E 2~\cite{dalle2}. Having shown that CLIP is memorizing individuals, suggests that those models could also leak information about people. As Carlini et al.~\cite{carlini_diffusion} have shown, it is possible to extract training images from these models, strongly suggesting that other information could be extracted as well. For example, naming individuals in these text-to-image models generates faithful images, allowing adversaries to possibly extract the appearance of a person just by naming them. 
Even though it is not as simple as just naming the person and getting an image back, our IDIA achieves something similar. In simple terms, the adversary is asking whether the model knows the person by using correct and incorrect predictions as a proxy for knowing and not knowing. 
As researchers are developing more and more capable (multi-modal) models, with even the goal of artificial general intelligence (AGI), future research has to investigate to which extent these models leak information about their training data. Consequently, the risks and benefits of increasing utility have to be weighed against user privacy concerns. Currently, many large-scale models are pushing the benchmarks simply by scaling the datasets. However, in most cases, more data means more knowledge. In particular, if a model can answer even complicated queries, an attacker might be able to extract sensitive information by asking the right questions.
Even though our experiments are not conclusive and future work has to investigate this further, we assume that there is a trade-off between the “emergent” abilities of machine learning models and their susceptibility to privacy attacks.

\subsection{A Tool to Strengthen User Privacy}
Usually, as with the CLIP~\cite{clip_radford} or DALL-E~\cite{dalle,dalle2} model from OpenAI, these large-scale scraped datasets are not publicly accessible, which is why individuals have no way of checking whether their data was unlawfully used for training. 
One way to verify this would be to request a copy of the user's data, which was processed by a company according to Article 15 of the GDPR~\cite{gdpr_eu}. However, as searching for similar images or texts in these huge datasets is very time-consuming and requires the data points to be manually inspected by humans to validate the data belonging to the requesting individual, it is very unlikely that companies and other providers of such models will give any disclosure whether the data was used. So far, it seems like no one has made such a request. Moreover, since the data is unstructured, it is even questionable what to look for in the data.

In the past, there have already been cases where companies were fined or had to delete their trained models because their model training was not compliant with data privacy laws.
For example, \textit{Clearview AI} is scraping billions of facial images to develop facial recognition software offered as a service, e.g., to law enforcement. As a result of violating the data protection laws in Canada, Australia, and parts of Europe, \textit{Clearview AI} has been fined by multiple countries~\cite{clearview_fine_uk,clearview_fine_italy,clearview_fine_greece}. 
Similarly, the company \textit{Everalbum}, provider of the now shut down cloud photo storage app “Ever”, was training facial recognition models on users' photos without their specific consent and was ordered by the US Federal Trade Commission to delete all models and algorithms trained on this user data~\cite{ever_techcrunch,ever_ftc}.
In a scenario where users only have access to the model, it is very difficult to verify the claims of the companies on what data the models were trained on. The setting where models were trained on data scraped from the internet is especially interesting, as data of people is used who have no relationship to the companies and might have never even heard of these firms or their products before. 

To verify companies' claims about their training data and to enforce privacy laws, we propose to use our IDIA to verify whether a person was used for training a model without consent. As shown in our experiments, the attack has a low false-positive and high true-positive rate, resulting in very high reliability and credibility of the results. This could make IDIAs especially interesting to use as evidence in court cases.
With this tool at hand, users could make a case against these corporations, demand their data to be removed from the training data and excluded from future model training, allowing them to enforce their “right to be forgotten” (according to GDPR~\cite{gdpr_eu}). \\
A demo of our IDIA for everyone to test if they have been used to train a model can be found at our Hugging Face Space\footnote{\url{https://huggingface.co/spaces/AIML-TUDA/does-clip-know-my-face}}.
 
\subsection{Possible Countermeasures}
Our IDIA exploits the fact that the model is memorizing the visual appearance of a person together with their name. A possible countermeasure to prevent these privacy attacks could be to mask out all names in the dataset, similar to the approach applied during the creation of the Conceptual Captions 3M dataset~\cite{cc3m}. 
However, it is questionable whether this will fully prevent IDIAs. 
If a person's name is memorized, other attributes about this individual are most likely memorized too. 
One could think of other textual attributes than the name of a person. Talking about celebrities, for example, one could think of films or TV series in which they played a role. For private individuals who post images on the internet, other possible attributes could be the occupation, the place of residency, or the usernames of social media accounts. 
Instead of using the person's name, one could imagine using these attributes to infer whether the person was in the training data.
With models getting more powerful, one should consider whether some decency could be taught to these multi-modal models. 
If the model is aware that the information it is about to share is private, e.g., by labeling sensitive data points, it could refuse to answer, similar to a doctor refusing to talk about the health status of other patients. However, since different people have different understandings of privacy, future work would have to investigate how one could create a general basis for privacy in machine learning models. 

\subsection{Social Impact and Ethical Considerations}
In our experiments, we have shown that CLIP memorizes sensitive data, i.e., facial appearance and the names of individuals. To perform our experiments in accordance with ethical guidelines, we only used images and names of celebrities and individuals of public interest and no data of private individuals was used. 
Even though our IDIA could be used to attack currently deployed systems, we believe it is important to share our results to foster a well-founded discussion about privacy in multi-modal models. Not only can IDIAs, as previously discussed, be used as a possible tool to strengthen user privacy, but they can also be used for future research as a measurement of how much information a model is leaking. As user privacy is currently in no way considered in the development and deployment of models like CLIP~\cite{clip_radford}, Stable Diffusion~\cite{rombach_diffusion} or DALL-E~\cite{dalle,dalle2}, we believe that it is essential to start a discussion about the trade-offs that are currently made between advancements in the AI community and the privacy of individuals.

\subsection{Future Work}
In our work, we have shown that it is possible to infer whether a person was used to train a vision-language model. However, we can imagine that a very similar technique could be used to infer the name of an individual on a picture. Future work could investigate whether these vision-language zero-shot classifiers could be misused as a model to identify individuals who were part of the training data. One possible way could be to use many possible first and last names and analyze how often the first and last names are predicted for each of the images. 
Another very interesting avenue for future work could be to develop a countermeasure. 
With privacy laws and regulations getting more important in the era of large machine learning models, the research area of unlearning is becoming increasingly important. Future research could investigate whether it is possible to unlearn an individual's data from these models after identifying them as part of the training data using the IDIA.

\section{Conclusion}\label{sec:conclusion}
In this paper, we have introduced a new type of privacy attack on multi-modal vision-language models, called Identity Inference Attack (IDIA), to assess information leakage of trained models. Through several large-scale experiments, we have demonstrated that with closed-box access to the model, a set of images and the name of individuals, an adversary can predict with very high accuracy whether facial images of a person were in the training data. Due to a low false-positive rate, the attack can be used by individuals to check whether their images were used to train a model without their consent. 
Specifically, our evaluation of models trained on the LAION-400M dataset has shown that even models trained on 400 million data points, 
still leak information about individuals only appearing less than 25 times in the dataset. This demonstrates that by posing the right questions, sensitive information about the training data of multi-modal vision-language models can be extracted, suggesting that there is a trade-off between the “emergent” abilities of a model and the privacy of training data. Overall, our results show that vision-language models indeed leak sensitive information about their training data and that privacy should be given greater consideration when training large-scale vision-language models.

\acks{The authors thank Daniel Neider for the fruitful discussions. This research has benefited from the Federal Ministry of Education and Research (BMBF) project KISTRA (reference no. 13N15343), the Hessian Ministry of Higher Education, Research, Science and the Arts (HMWK) cluster projects “The Third Wave of AI” and hessian.AI, from the German Center for Artificial Intelligence (DFKI) project “SAINT,” as well as from the joint ATHENE project of the HMWK and the BMBF “AVSV”.
}

\renewcommand{\theHsection}{A\arabic{section}}
\appendix
\onecolumn
\section{Hard- and Software Details}
The experiments conducted in this work were run on NVIDIA DGX machines with NVIDIA DGX Server Version 5.1.0 and Ubuntu 20.04.4 LTS. The machines have NVIDIA A100-SXM4-40GB GPUs, AMD EPYC 7742 64-Core processors and 1.9TB of RAM. The experiments were run with Python 3.8, CUDA 11.3 and PyTorch 1.12.1 with TorchVision 0.13.1.

\section{Prompt Templates and Generation of Possible Names}\label{app:prompt_temp_name_gen}
The prompt templates used for our IDIA can be seen in Table~\ref{tab:prompt_templates}.
In each of the prompt templates, \textit{X} was substituted with possible names. The possible names were generated by combining the most popular male and female first names in the US from 1880-2008~\cite{first_names_list} equally often with the most often last names in the US~\cite{last_names_list}. Out of all combinations, we randomly sampled $1000$ names for our experiments.
\begin{table}[H]
    \centering
    \resizebox{\textwidth}{!}{
        \begin{tabular}{lll}
        \hline
        \multicolumn{3}{c}{Prompt Templates}\\
        \hline
        X                                       & a woman named \textit{X}                      & a colored photo of \textit{X}         \\
        an image of \textit{X}                  & the name of the person is \textit{X}          & a black and white photo of \textit{X} \\
        a photo of \textit{X}                   & a photo of a person with the name \textit{X}  & a cool photo of \textit{X}            \\
        \textit{X} on a photo                   & \textit{X} at a gala                          & a cropped photo of \textit{X}         \\
        a photo of a person named \textit{X}    & a photo of the celebrity \textit{X}           & a cropped image of \textit{X}         \\
        a person named \textit{X}               & actor \textit{X}                              & \textit{X} in a suite                 \\
        a man named \textit{X}                  & actress \textit{X}                            & \textit{X} in a dress                 
        \end{tabular}
    }
    \caption{Prompt templates used for the IDIA in the experiments. \textit{X} was substituted with the names of the individuals.}
    \label{tab:prompt_templates}
\end{table}

\section{Analysis of the CC3M Dataset}\label{app:cc3m_analysis}
While creating the CC3M dataset, Sharma et al.~\cite{cc3m} have anonymized entity names in the image captions. 
Therefore, it is not possible to search for the individual names in the image captions. 
To still be able to analyze which individuals are already present in the dataset, we trained a ResNet-50 on the FaceScrub dataset to classify the faces of the 530 celebrities used for the experiments.
The ResNet-50 was trained with a batch size of 128 for 100 epochs on a single A100 GPU using the cropped images of the FaceScrub dataset. 
90\% of the dataset was used to train the model, while the other 10\% was used as a validation set. 
All images were resized to 224x224 pixels and center cropped. 
During training, randomly resized cropping, color jitter, and horizontal flipping were applied as data augmentation techniques.
We used Adam~\cite{adam} with a learning rate of $0.003$, which was reduced by $0.1$ after 75 and 90 epochs.
The trained ResNet-50 achieved a test accuracy of 91\% on the FaceScrub test set. 
We then used clip-retrieval~\cite{clip_retrieval} to retrieve similar images for each of the celebrities in the FaceScrub dataset from the CC3M dataset.
OpenCV's~\cite{opencv_library} Haar cascade face detector was then used to detect faces in those similar images and the trained ResNet-50 was used to predict whether the person in question was present in the images. 
We then chose the 200 individuals with the lowest number of correct predictions and mixed the first 100 of these individuals with the CC3M dataset, while the other half of the individuals were not used for training the target models of our experiments. Even though no pictures of people with the corresponding names are in the dataset due to anonymization when the dataset was created, this procedure reduces the probability that pictures of individuals used for the experiments are present in the dataset at all.

\section{Target Models Trained on CC3M}\label{app:target_models}
To train the CLIP models on the CC3M dataset, we used the code of the open-source project OpenCLIP~\cite{open_clip}.
As can be seen in Table~\ref{app:clip_cc_training_results}, we trained CLIP models with three different vision model architectures. 
The hyperparameters of the vision and text model architectures are the same as in the original CLIP paper~\cite{clip_radford}. 
All the models were trained on 8 A100 GPUs for 50 epochs on the CC3M dataset with a per GPU batch size of 128, a learning rate of $1e^{-3}$, and a weight decay parameter of $0.1$. 
The top-1 and top-5 ImageNet zero-shot prediction accuracy of the trained models can be seen in Table~\ref{app:clip_cc_training_results}. 
While the two ResNets achieve comparable results of a top-1 accuracy of around 20\%, the vision transformer ViT-B/32 achieves a top-1 accuracy of 14.3\% on the ImageNet validation set~\shortcite{imagenet}. 
The image-to-text rank of the models describes how often the model predicts the correct text prompt for a given validation image. Rank@1 means that the ground truth caption for the image was the predicted one, while rank@5 means that the ground truth predicted caption was under the top-5 predicted ones. 
As with the zero-shot accuracy, one can see that the ResNet-50 and the ResNet-50x4 achieve comparable results. While the ResNet-50 achieves an image-to-text rank@1 of around 30\%, the ResNet-50x4 achieves an image-to-text rank@1 of around 33\%, meaning that in about a third of the cases, the correct caption to a given image in the validation set was predicted.
\begin{table*}[ht]
    \centering
    \resizebox{\linewidth}{!}{
    \begin{tabular}{lcccccc}
    \toprule
    Architecture                & Number of Image-Text Pairs per Person & ImageNet Val Top-1 Acc    & ImageNet Val Top-5 Acc    & Val Image-to-Text Rank@1  & Val Image-to-Text Rank@5      \\
    \midrule
    \multirow{4}{*}{ViT-B/32}   & 75                                    & 14.3\%                    & 29.8\%                    & 21.2\%                    & 39.9\%                        \\
                                & 50                                    & 14.3\%                    & 29.8\%                    & 21.0\%                    & 39.7\%                        \\
                                & 25                                    & 14.7\%                    & 30.1\%                    & 21.2\%                    & 39.9\%                        \\
                                & 10                                    & 14.5\%                    & 30.0\%                    & 21.1\%                    & 39.8\%                        \\
                                & 5                                     & 14.5\%                    & 29.5\%                    & 21.0\%                    & 39.6\%                        \\
                                & 1                                     & 14.3\%                    & 29.6\%                    & 21.2\%                    & 39.6\%                        \\ 
    \hline
    \multirow{4}{*}{ResNet-50}  & 75                                    & 19.6\%                    & 37.4\%                    & 30.9\%                    & 53.4\%                        \\
                                & 50                                    & 19.9\%                    & 37.9\%                    & 30.9\%                    & 53.4\%                        \\
                                & 25                                    & 20.0\%                    & 38.0\%                    & 31.3\%                    & 53.4\%                        \\
                                & 10                                    & 19.7\%                    & 38.0\%                    & 31.3\%                    & 53.6\%                        \\
                                & 5                                     & 19.7\%                    & 38.1\%                    & 31.1\%                    & 53.6\%                        \\
                                & 1                                     & 19.3\%                    & 38.2\%                    & 30.9\%                    & 52.7\%                        \\ 
    \hline
    \multirow{4}{*}{ResNet-50x4}& 75                                    & 21.0\%                    & 39.8\%                    & 33.7\%                    & 56.4\%                        \\
                                & 50                                    & 21.7\%                    & 40.2\%                    & 33.6\%                    & 56.2\%                        \\
                                & 25                                    & 20.7\%                    & 39.5\%                    & 33.1\%                    & 56.2\%                        \\
                                & 10                                    & 21.2\%                    & 39.9\%                    & 33.4\%                    & 55.7\%                        \\
                                & 5                                     & 21.0\%                    & 39.3\%                    & 33.8\%                    & 56.2\%                        \\
                                & 1                                     & 21.3\%                    & 39.9\%                    & 33.5\%                    & 55.8\%                        \\ 
    \bottomrule
    \end{tabular}
    }
    \caption{CLIP performs well on zero-shot predictions, here trained on the CC3M dataset, and does not overfit. A random classifier would achieve a top-1 accuracy (the higher, the better) of $0.1\%$ as there are $1000$ classes in ImageNet. Image-to-Text Rank (the higher, the better) is the fraction of images where the rank is zero, meaning that the image and the according caption are indeed the most similar in the dataset. Moreover, the number of image-text pairs added to the CC3M training set has no influence on the predictive performance, showing that they do not overfit to individuals.}
    \label{app:clip_cc_training_results}
\end{table*}

\section{Additional Experimental Results}\label{app:additional_experimental_results}
\begin{figure*}[ht]
    \centering
    \includegraphics[width=.5\textwidth]{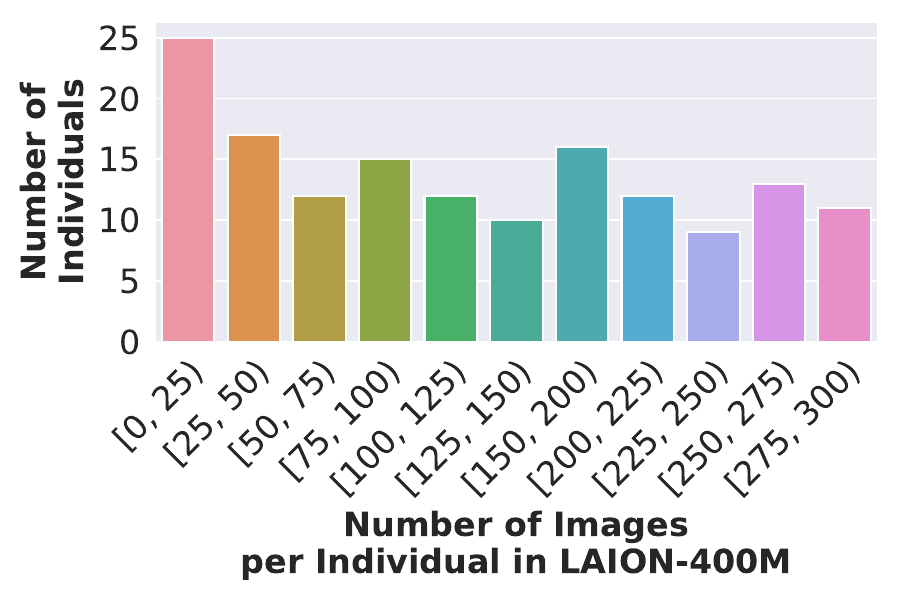}
    \caption{Histogram of how many individuals used for the experiments with the LAION-400M dataset are in each bin.}
    \label{fig:num_training_samples_hist}
\end{figure*}

\begin{figure*}[ht]
    \centering
    \resizebox{.9\textwidth}{!}{%
        \begin{subfigure}[b]{0.32\textwidth}
            \centering
            \includegraphics[width=\textwidth]{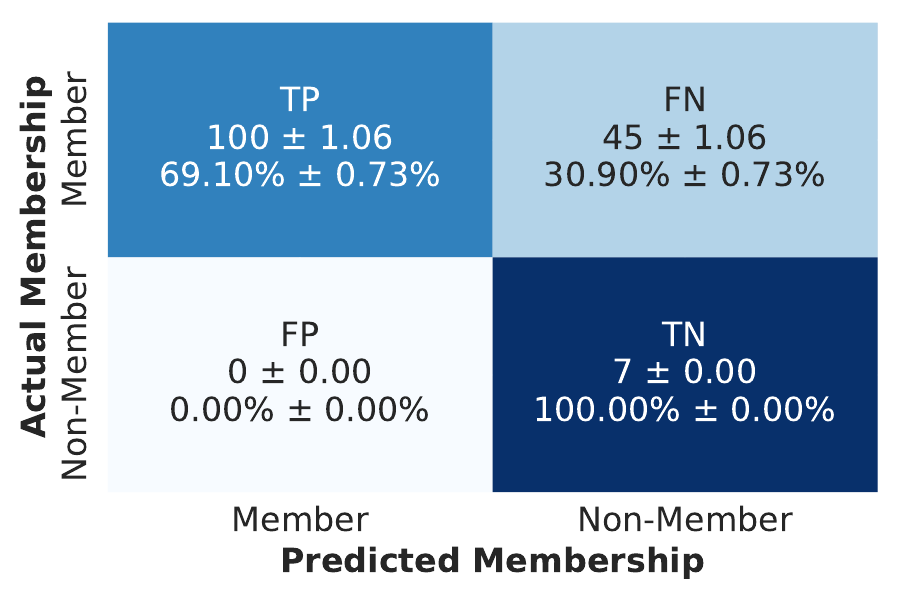}
            \caption{LAION-400M ViT-B/32}
        \end{subfigure}
        \begin{subfigure}[b]{0.32\textwidth}
            \centering
            \includegraphics[width=\textwidth]{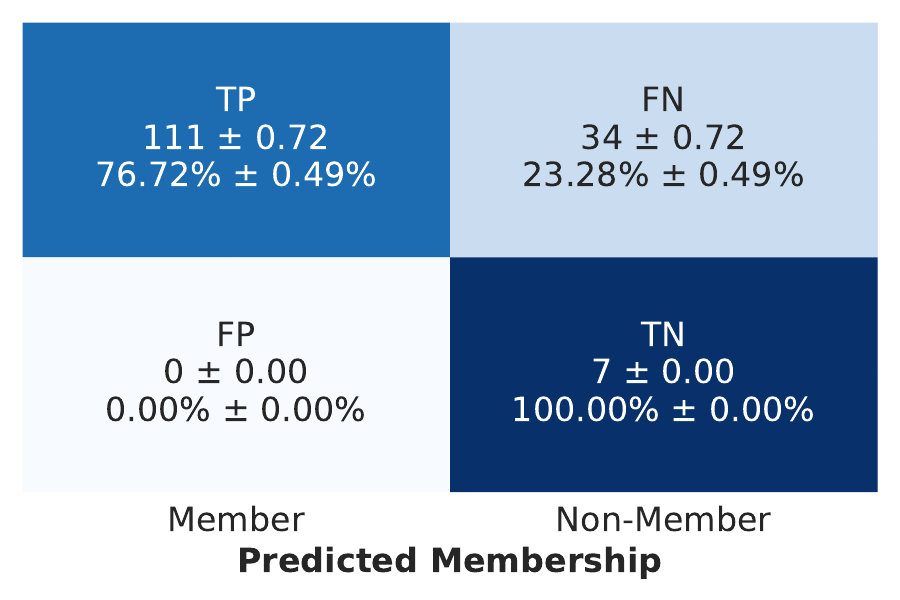}
            \caption{LAION-400M ViT-B/16}
        \end{subfigure}
        \begin{subfigure}[b]{0.32\textwidth}
            \centering
            \includegraphics[width=\textwidth]{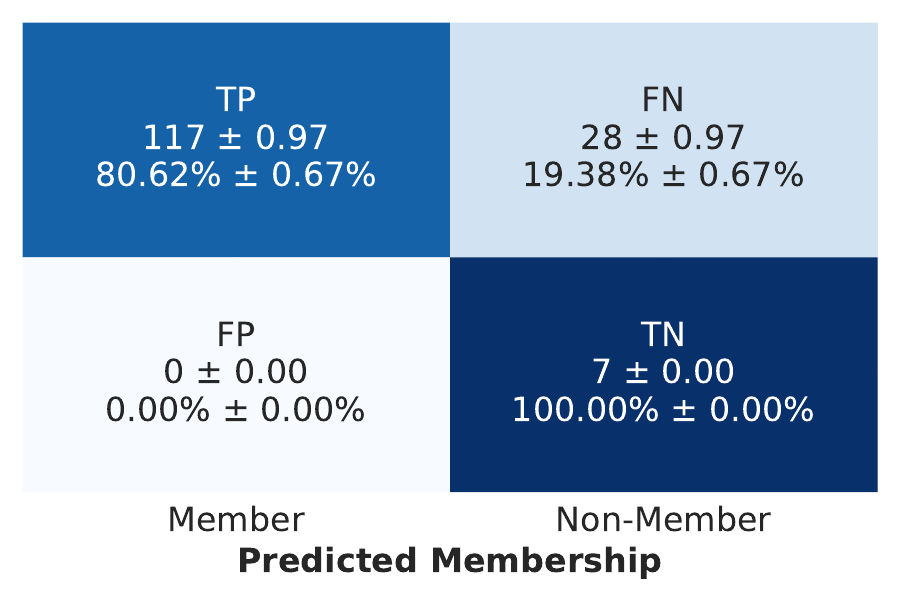}
            \caption{LAION-400M ViT-L/14}
        \end{subfigure}
    }
    \resizebox{.9\textwidth}{!}{
        \begin{subfigure}[b]{0.32\textwidth}
            \centering
            \includegraphics[width=\textwidth]{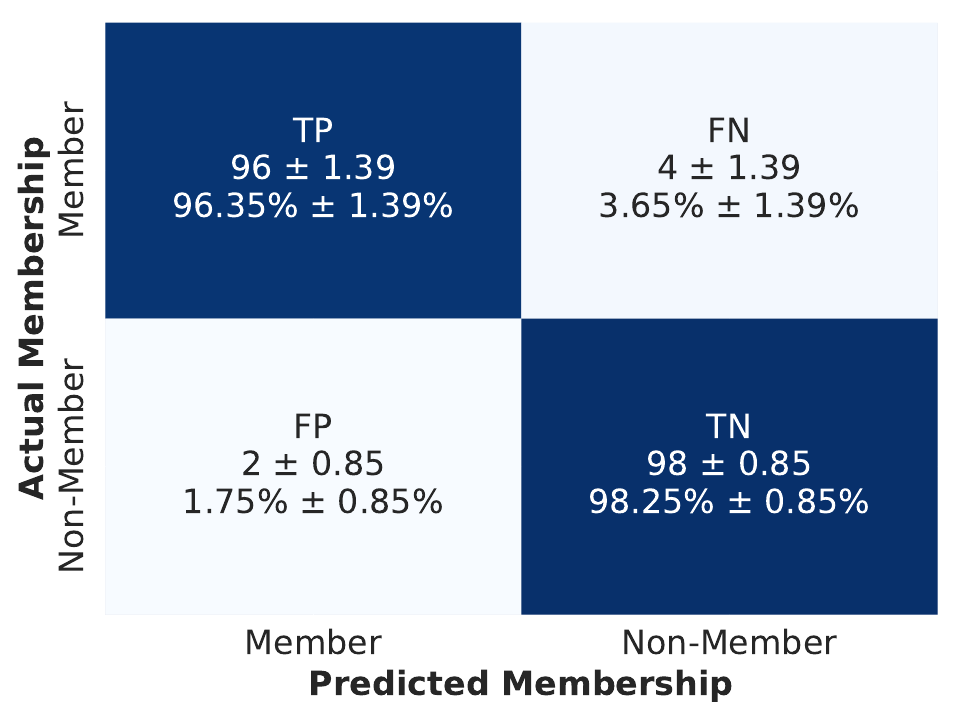}
            \caption{CC3M RN50}
        \end{subfigure}
        \begin{subfigure}[b]{0.32\textwidth}
            \centering
            \includegraphics[width=\textwidth]{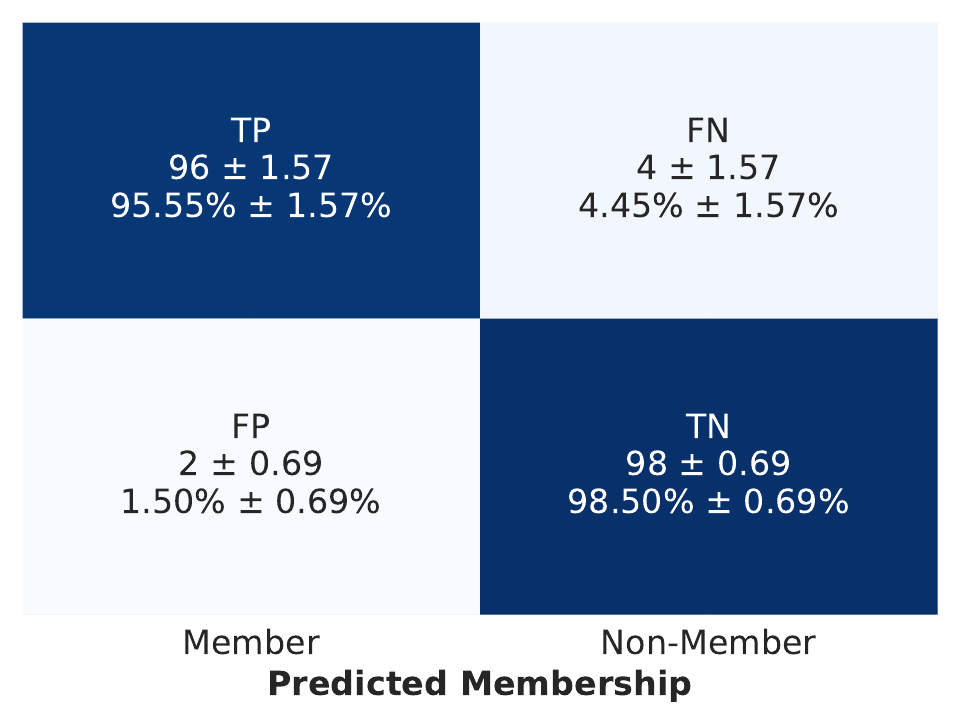}
            \caption{CC3M RN50x4}
        \end{subfigure}
        \begin{subfigure}[b]{0.32\textwidth}
            \centering
            \includegraphics[width=\textwidth]{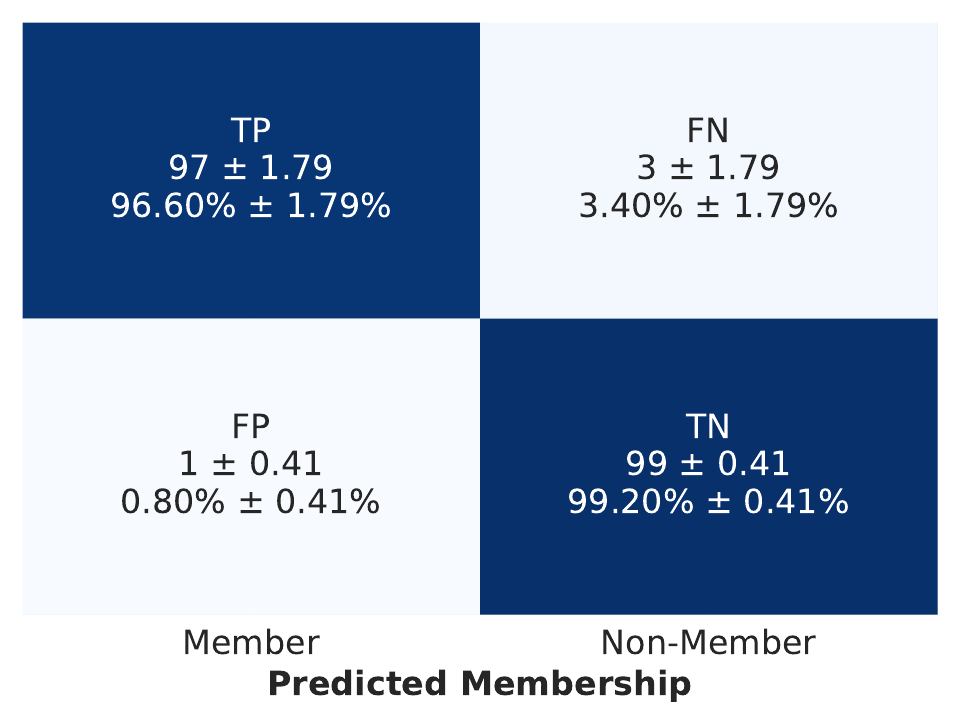}
            \caption{CC3M ViT-B/32}
        \end{subfigure}
    }
    \caption{Confusion matrices for the IDIA using 30 attack images, performed on the models trained on LAION-400M and CC3M.}
    \label{fig:confusion_matrices}
\end{figure*}

\begin{figure*}[hb]
    \centering
    \resizebox{.9\textwidth}{!}{
        \begin{subfigure}[b]{0.34\textwidth}
            \centering
            \includegraphics[width=\textwidth]{plots/num_training_samples/num_training_samples_plot_multiprompt_laion400_ViT-L-14_1000_300_1.pdf}
            \caption{LAION-400M ViT-L/14}
        \end{subfigure}
        \begin{subfigure}[b]{0.325\textwidth}
            \centering
            \includegraphics[width=\textwidth]{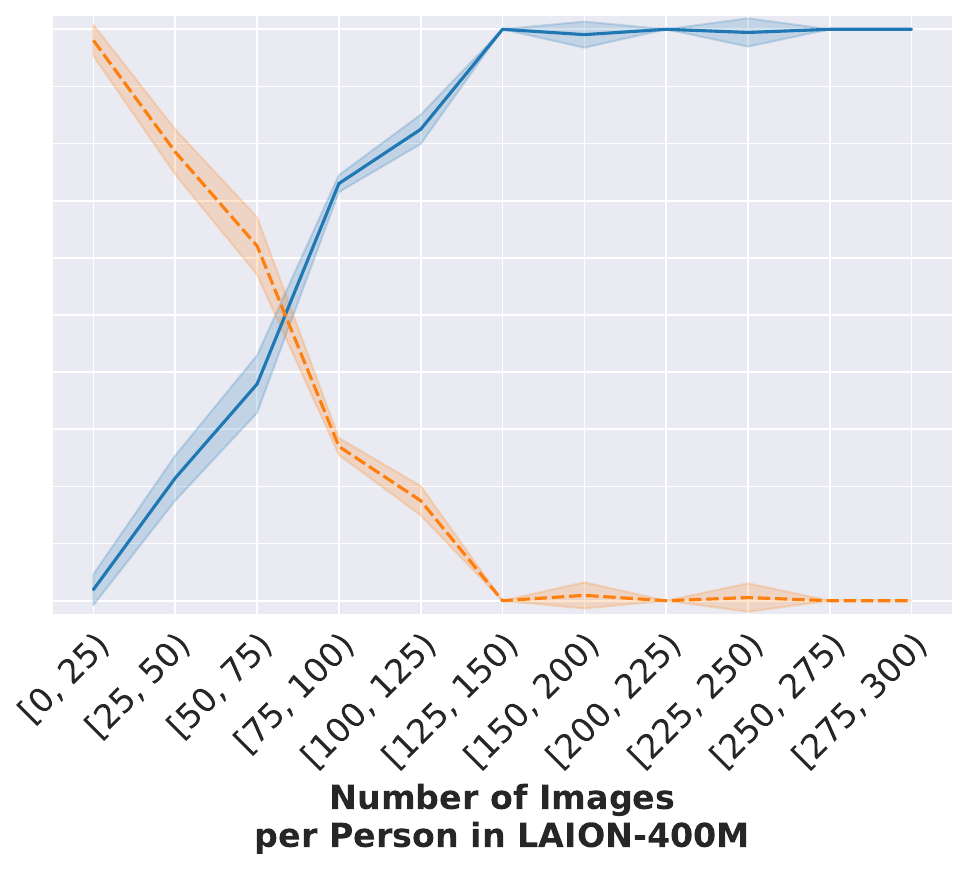}
            \caption{LAION-400M ViT-B/32}
        \end{subfigure}
        \begin{subfigure}[b]{0.325\textwidth}
            \centering
            \includegraphics[width=\textwidth]{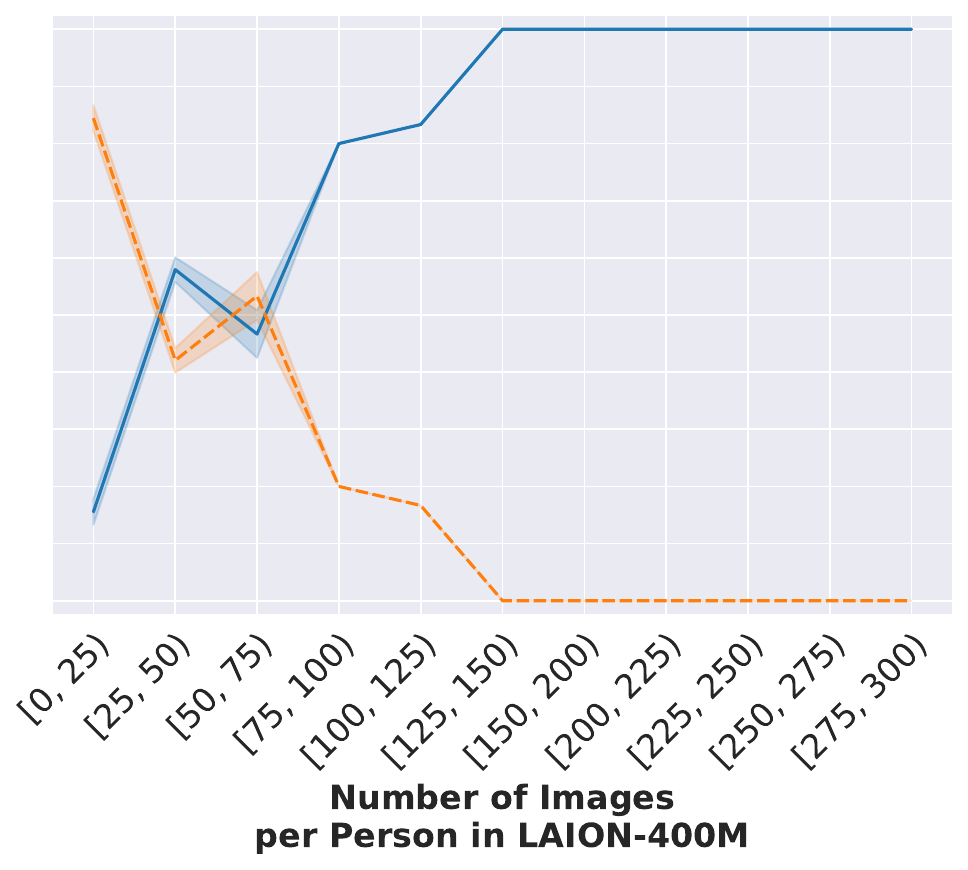}
            \caption{LAION-400M ViT-B/16}
        \end{subfigure}
    }
    \resizebox{.9\textwidth}{!}{
        \begin{subfigure}[b]{0.345\textwidth}
            \centering
            \includegraphics[width=\textwidth]{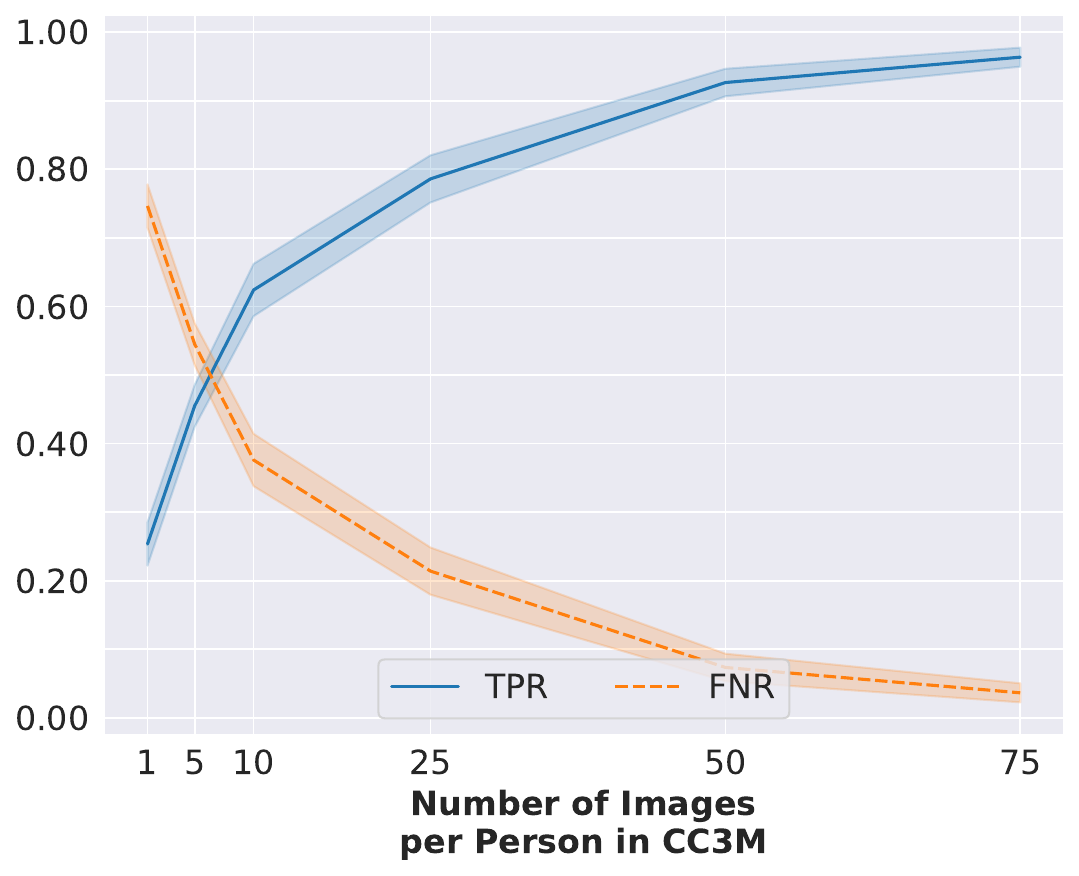}
            \caption{CC3M RN50}
        \end{subfigure}
        \begin{subfigure}[b]{0.315\textwidth}
            \centering
            \includegraphics[width=\textwidth]{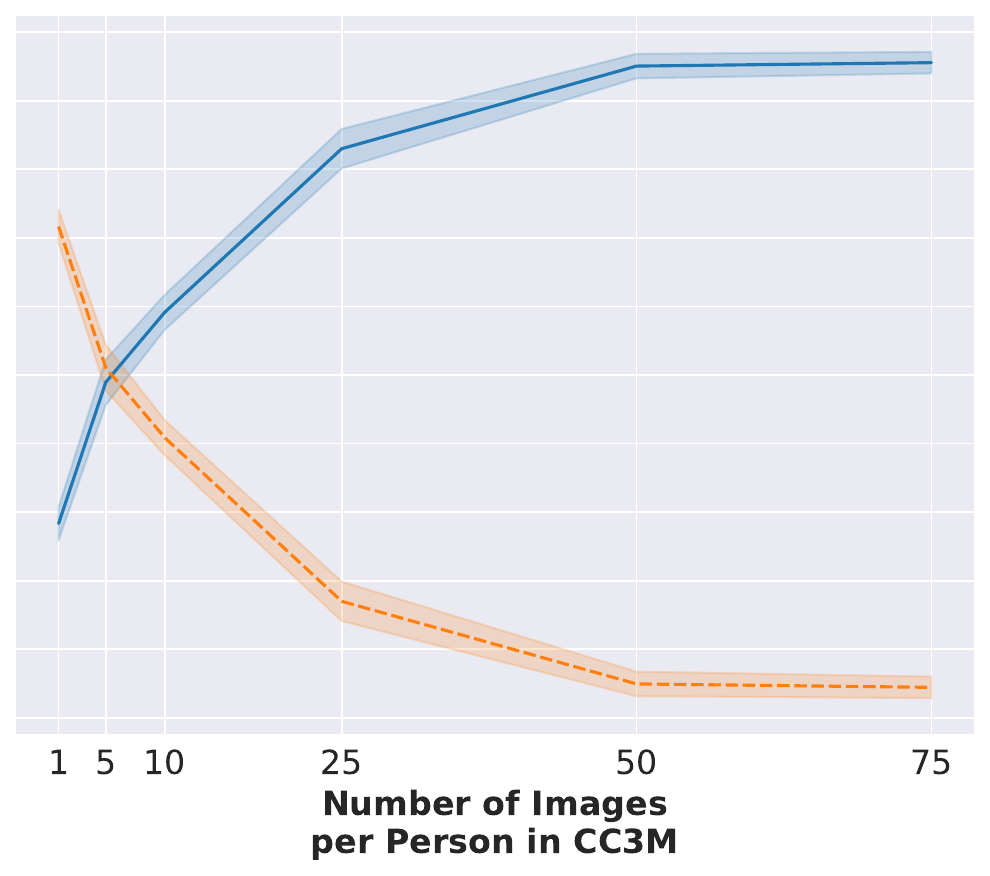}
            \caption{CC3M RN50x4}
        \end{subfigure}
        \begin{subfigure}[b]{0.315\textwidth}
            \centering
            \includegraphics[width=\textwidth]{plots/num_training_samples/num_training_samples_plot_multiprompt_CC2M_ViT-B-32_1000_1.pdf}
            \caption{CC3M ViT-B/32}
        \end{subfigure}
    }
    \caption{Depicted are the mean and standard deviation of the true-positive rate and false-negative rate of the IDIA for different number of training images per person. In this experiment 30 samples were used to perform the attack.}
    \label{fig:add_num_training_sample_plots}
\end{figure*}

\begin{figure}[ht]
    \centering
    \begin{subfigure}[b]{0.335\textwidth}
        \centering
        \includegraphics[width=\textwidth]{plots/heatmaps/heatmap_num_training_samples_laion400_ViT-B-32_1000_300_1.pdf}
        \caption{LAION-400M ViT-B/32}
    \end{subfigure}
    \begin{subfigure}[b]{0.324\textwidth}
        \centering
        \includegraphics[width=\textwidth]{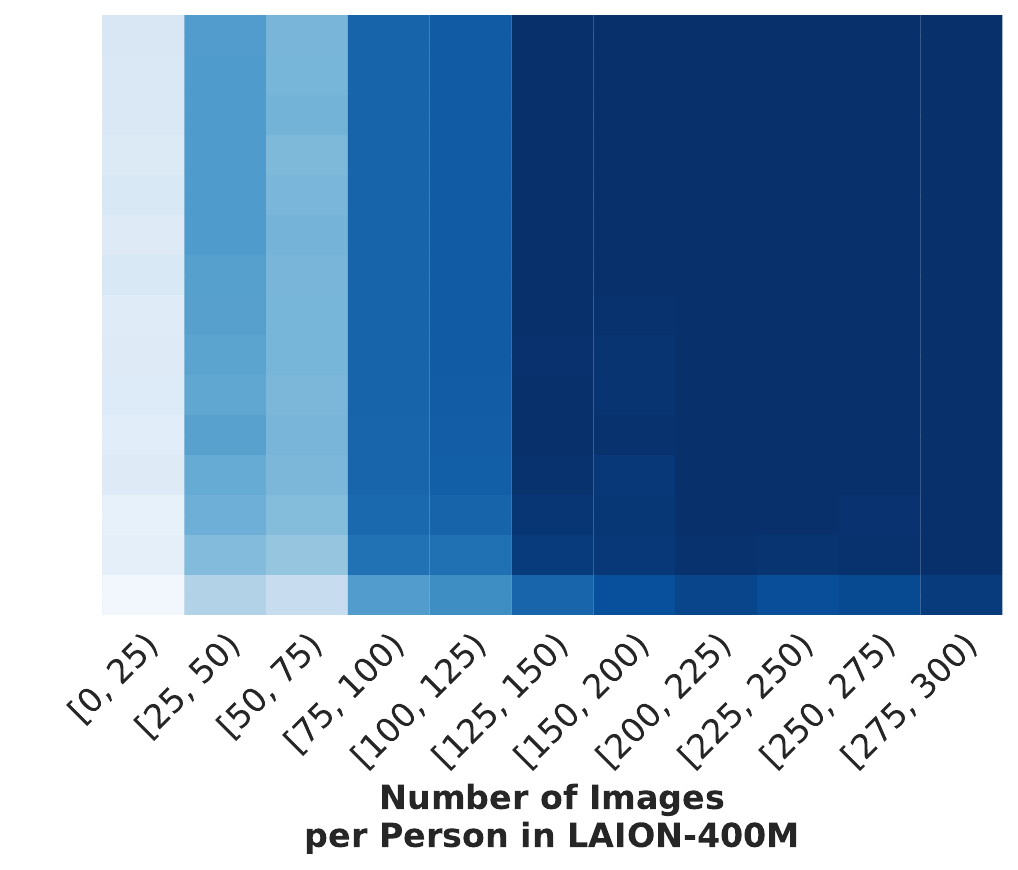}
        \caption{LAION-400M ViT-B/16}
    \end{subfigure}
    \begin{subfigure}[b]{0.324\textwidth}
        \centering
        \includegraphics[width=\textwidth]{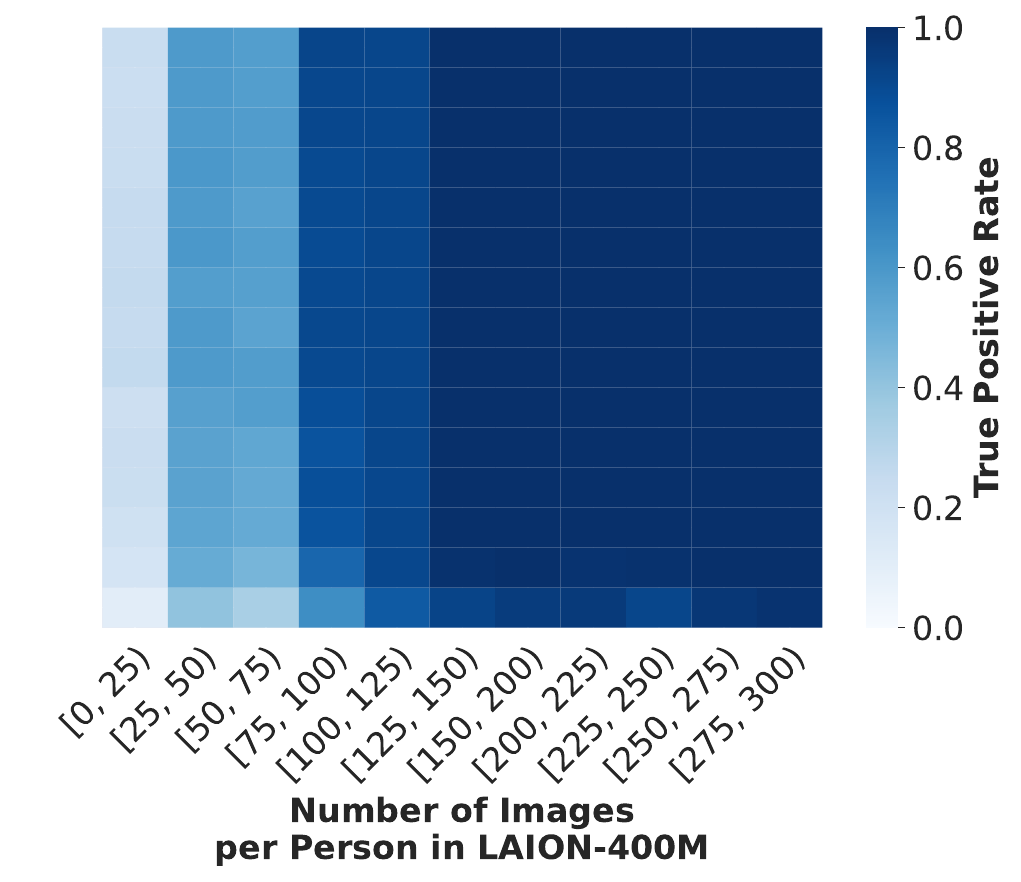}
        \caption{LAION-400M ViT-L/14}
    \end{subfigure}
    \begin{subfigure}[b]{0.32\textwidth}
        \centering
        \includegraphics[width=\textwidth]{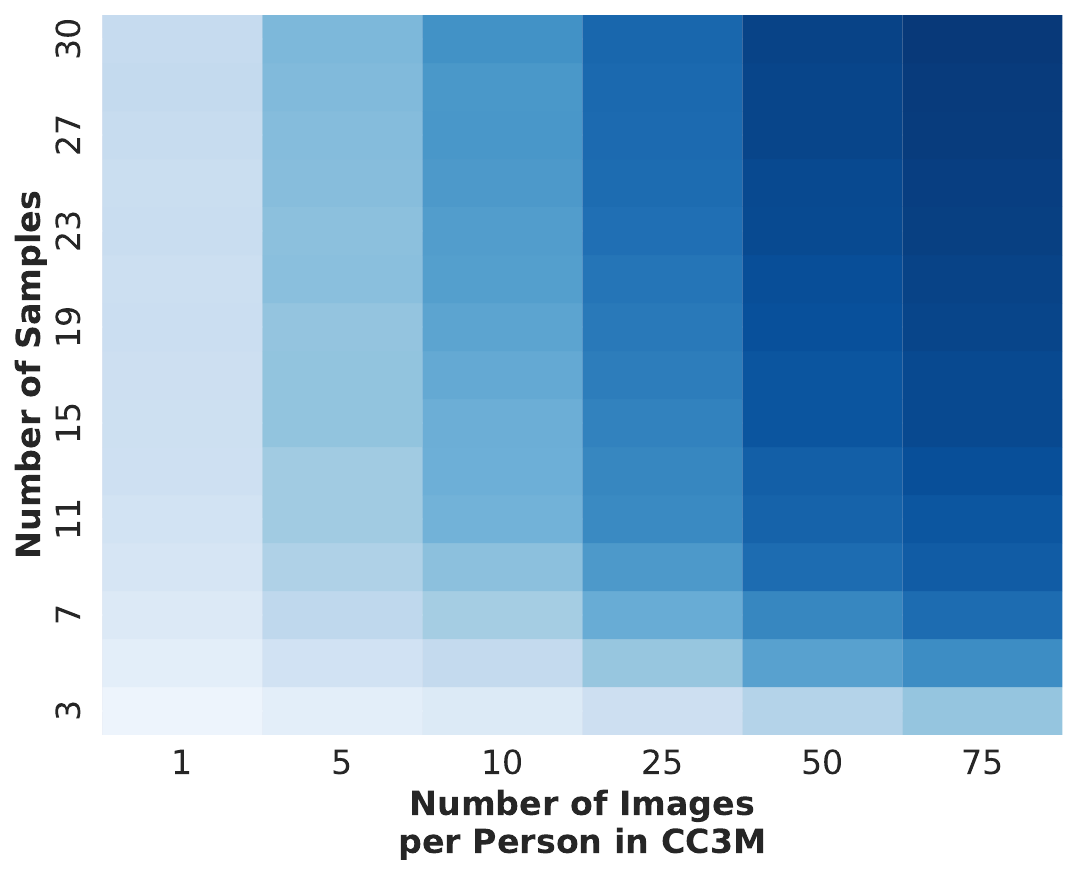}
        \caption{CC3M RN50}
    \end{subfigure}
    \begin{subfigure}[b]{0.32\textwidth}
        \centering
        \includegraphics[width=\textwidth]{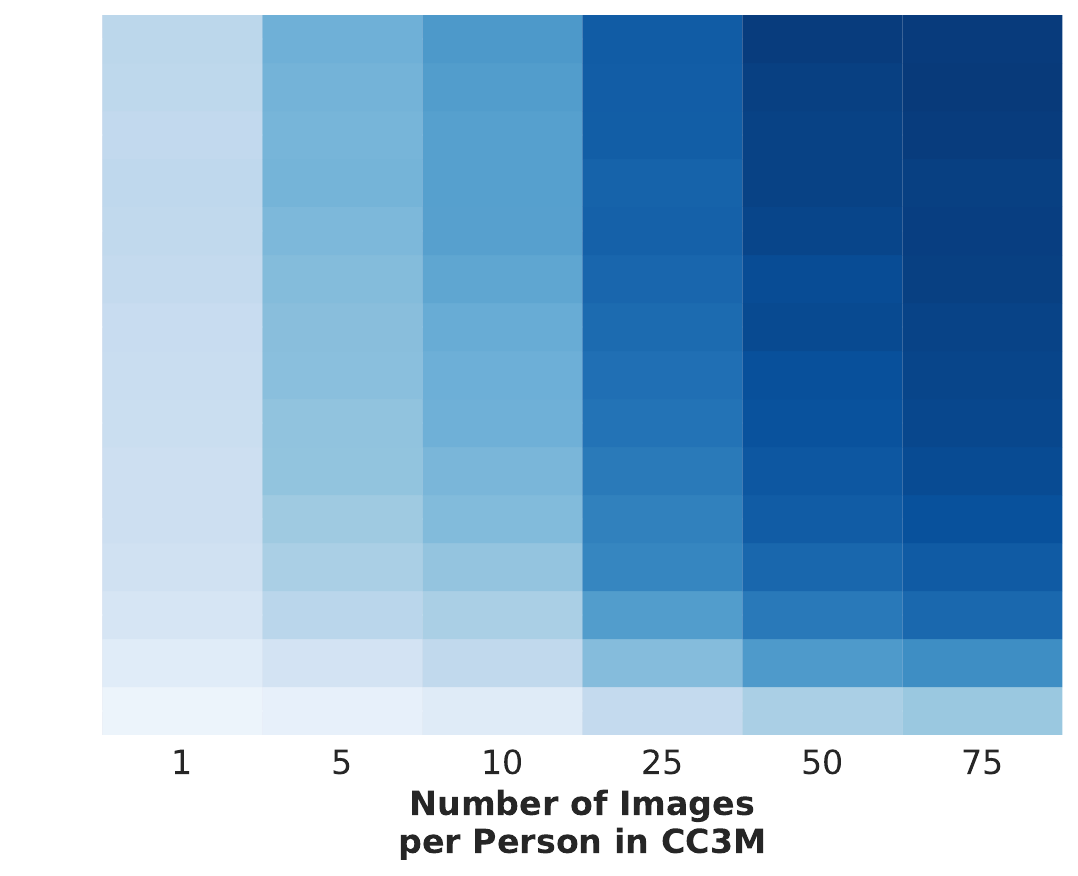}
        \caption{CC3M RN50x4}
    \end{subfigure}
    \begin{subfigure}[b]{0.32\textwidth}
        \centering
        \includegraphics[width=\textwidth]{plots/heatmaps/heatmap_num_training_samples_CC2M_ViT-B-32_1000_1.pdf}
        \caption{CC3M ViT-B/32}
    \end{subfigure}
    \caption{Influence of the number of samples available to the attacker and the number of samples per person in the training data on the true-positive rate (TPR) of the attack. With a higher number of training samples per individual, the advantage of having more samples available for the attack is less pronounced for each of the models.}
    \label{fig:add_heatmaps}
\end{figure}

The number of individuals for each of the bins of different occurrences in the LAION-400M dataset can be seen in Fig.~\ref{fig:num_training_samples_hist}. The confusion matrices of the IDIA using 30 attack samples on the LAION and the CC3M dataset can be seen in Fig.~\ref{fig:confusion_matrices}. The additional plots of the performance of IDIA depending on how many samples were used for training the models can be seen in Fig.~\ref{fig:add_num_training_sample_plots}. Additional heatmaps of the experiments can be seen in Fig.~\ref{fig:add_heatmaps}.

\clearpage
\vskip 0.2in
\bibliography{main}

\begin{thebibliography}{}

\bibitem[\protect\BCAY{Alayrac, Donahue, Luc, Miech, Barr, Hasson, Lenc,
  Mensch, Millican, Reynolds, Ring, Rutherford, Cabi, Han, Gong, Samangooei,
  Monteiro, Menick, Borgeaud, Brock, Nematzadeh, Sharifzadeh, Bi\'{n}kowski,
  Barreira, Vinyals, Zisserman,\ \BBA\ Simonyan}{Alayrac
  et~al.}{2022}]{flamingo}
Alayrac, J.-B., Donahue, J., Luc, P., Miech, A., Barr, I., Hasson, Y., Lenc,
  K., Mensch, A., Millican, K., Reynolds, M., Ring, R., Rutherford, E., Cabi,
  S., Han, T., Gong, Z., Samangooei, S., Monteiro, M., Menick, J.~L., Borgeaud,
  S., Brock, A., Nematzadeh, A., Sharifzadeh, S., Bi\'{n}kowski, M.~a.,
  Barreira, R., Vinyals, O., Zisserman, A., \BBA\ Simonyan, K. \BBOP2022\BBCP.
\newblock \BBOQ {Flamingo: a Visual Language Model for Few-Shot Learning}\BBCQ\
\newblock In {\Bem Advances in Neural Information Processing Systems
  (NeurIPS)}.

\bibitem[\protect\BCAY{Beaumont}{Beaumont}{2022}]{clip_retrieval}
Beaumont, R. \BBOP2022\BBCP.
\newblock \BBOQ {Clip Retrieval: Easily compute clip embeddings and build a
  clip retrieval system with them}\BBCQ\
\newblock GitHub. \url{https://github.com/rom1504/clip-retrieval}.

\bibitem[\protect\BCAY{Bradski}{Bradski}{2000}]{opencv_library}
Bradski, G. \BBOP2000\BBCP.
\newblock \BBOQ {The OpenCV Library}\BBCQ\
\newblock In {\Bem Dr. Dobb's Journal of Software Tools}.

\bibitem[\protect\BCAY{Brown, Mann, Ryder, Subbiah, Kaplan, Dhariwal,
  Neelakantan, Shyam, Sastry, Askell, Agarwal, Herbert-Voss, Krueger, Henighan,
  Child, Ramesh, Ziegler, Wu, Winter, Hesse, Chen, Sigler, Litwin, Gray, Chess,
  Clark, Berner, McCandlish, Radford, Sutskever,\ \BBA\ Amodei}{Brown
  et~al.}{2020}]{gpt3}
Brown, T., Mann, B., Ryder, N., Subbiah, M., Kaplan, J.~D., Dhariwal, P.,
  Neelakantan, A., Shyam, P., Sastry, G., Askell, A., Agarwal, S.,
  Herbert-Voss, A., Krueger, G., Henighan, T., Child, R., Ramesh, A., Ziegler,
  D., Wu, J., Winter, C., Hesse, C., Chen, M., Sigler, E., Litwin, M., Gray,
  S., Chess, B., Clark, J., Berner, C., McCandlish, S., Radford, A., Sutskever,
  I., \BBA\ Amodei, D. \BBOP2020\BBCP.
\newblock \BBOQ {Language Models are Few-Shot Learners}\BBCQ\
\newblock In {\Bem Advances in Neural Information Processing Systems
  (NeurIPS)}.

\bibitem[\protect\BCAY{Carlini, Chien, Nasr, Song, Terzis,\ \BBA\
  Tram{\`{e}}r}{Carlini et~al.}{2022}]{carliniMIA}
Carlini, N., Chien, S., Nasr, M., Song, S., Terzis, A., \BBA\ Tram{\`{e}}r, F.
  \BBOP2022\BBCP.
\newblock \BBOQ {Membership Inference Attacks From First Principles}\BBCQ\
\newblock In {\Bem IEEE Symposium on Security and Privacy (S\&P)}.

\bibitem[\protect\BCAY{Carlini, Hayes, Nasr, Jagielski, Sehwag, Tram{\`e}r,
  Balle, Ippolito,\ \BBA\ Wallace}{Carlini et~al.}{2023}]{carlini_diffusion}
Carlini, N., Hayes, J., Nasr, M., Jagielski, M., Sehwag, V., Tram{\`e}r, F.,
  Balle, B., Ippolito, D., \BBA\ Wallace, E. \BBOP2023\BBCP.
\newblock \BBOQ {Extracting Training Data from Diffusion Models}\BBCQ\
\newblock In {\Bem USENIX Security Symposium}.

\bibitem[\protect\BCAY{Cherti, Beaumont, Wightman, Wortsman, Ilharco, Gordon,
  Schuhmann, Schmidt,\ \BBA\ Jitsev}{Cherti et~al.}{2023}]{mehdi_clip_scaling}
Cherti, M., Beaumont, R., Wightman, R., Wortsman, M., Ilharco, G., Gordon, C.,
  Schuhmann, C., Schmidt, L., \BBA\ Jitsev, J. \BBOP2023\BBCP.
\newblock \BBOQ {Reproducible Scaling Laws for Contrastive Language-Image
  Learning}\BBCQ\
\newblock In {\Bem IEEE/CVF Conference on Computer Vision and Pattern
  Recognition (CVPR)}.

\bibitem[\protect\BCAY{Choquette-Choo, Tramer, Carlini,\ \BBA\
  Papernot}{Choquette-Choo et~al.}{2021}]{choquette2021}
Choquette-Choo, C.~A., Tramer, F., Carlini, N., \BBA\ Papernot, N.
  \BBOP2021\BBCP.
\newblock \BBOQ {Label-Only Membership Inference Attacks}\BBCQ\
\newblock In {\Bem International Conference on Machine Learning (ICML)}.

\bibitem[\protect\BCAY{Dosovitskiy, Beyer, Kolesnikov, Weissenborn, Zhai,
  Unterthiner, Dehghani, Minderer, Heigold, Gelly, Uszkoreit,\ \BBA\
  Houlsby}{Dosovitskiy et~al.}{2021}]{vit}
Dosovitskiy, A., Beyer, L., Kolesnikov, A., Weissenborn, D., Zhai, X.,
  Unterthiner, T., Dehghani, M., Minderer, M., Heigold, G., Gelly, S.,
  Uszkoreit, J., \BBA\ Houlsby, N. \BBOP2021\BBCP.
\newblock \BBOQ {An Image is Worth 16x16 Words: Transformers for Image
  Recognition at Scale}\BBCQ\
\newblock In {\Bem International Conference on Learning Representations
  (ICLR)}.

\bibitem[\protect\BCAY{Edwards}{Edwards}{2022}]{arstechnica_medical_imgs_laion}
Edwards, B. \BBOP2022\BBCP.
\newblock \BBOQ {Artist finds private medical record photos in popular AI
  training data set}\BBCQ\
\newblock arsTechnica.
  \url{https://arstechnica.com/information-technology/2022/09/artist-finds-private-medical-record-photos-in-popular-ai-training-data-set/}.
\newblock Online; accessed 21-January-2023.

\bibitem[\protect\BCAY{Eichenberg, Black, Weinbach, Parcalabescu,\ \BBA\
  Frank}{Eichenberg et~al.}{2022}]{eichenberg2021}
Eichenberg, C., Black, S., Weinbach, S., Parcalabescu, L., \BBA\ Frank, A.
  \BBOP2022\BBCP.
\newblock \BBOQ {{MAGMA} {--} Multimodal Augmentation of Generative Models
  through Adapter-based Finetuning}\BBCQ\
\newblock In {\Bem Findings of the Association for Computational Linguistics:
  EMNLP 2022}.

\bibitem[\protect\BCAY{Estrada\ \BBA\ Walko}{Estrada\ \BBA\
  Walko}{2022}]{weight_watchers_ftc}
Estrada, D.\BBACOMMA\  \BBA\ Walko, D. \BBOP2022\BBCP.
\newblock \BBOQ {{FTC} Takes Action Against Company Formerly Known as Weight
  Watchers for Illegally Collecting Kids’ Sensitive Health Data}\BBCQ\
\newblock FTC.
  \url{https://www.ftc.gov/news-events/news/press-releases/2022/03/ftc-takes-action-against-company-formerly-known-weight-watchers-illegally-collecting-kids-sensitive}.
\newblock Online; accessed 27-January-2023.

\bibitem[\protect\BCAY{{European Parliament and European Council}}{{European
  Parliament and European Council}}{2016}]{gdpr_eu}
{European Parliament and European Council} \BBOP2016\BBCP.
\newblock \BBOQ {Regulation (EU) 2016/679 of the European Parliament and of the
  Council of 27 April 2016 on the protection of natural persons with regard to
  the processing of personal data and on the free movement of such data, and
  repealing Directive 95/46/EC (General Data Protection Regulation)}\BBCQ\
\newblock
  \url{https://eur-lex.europa.eu/legal-content/EN/TXT/?uri=celex\%3A32016R0679}.
\newblock Online; accessed 22-January-2023.

\bibitem[\protect\BCAY{Fredrikson, Jha,\ \BBA\ Ristenpart}{Fredrikson
  et~al.}{2015}]{fredrikson2015}
Fredrikson, M., Jha, S., \BBA\ Ristenpart, T. \BBOP2015\BBCP.
\newblock \BBOQ {Model Inversion Attacks That Exploit Confidence Information
  and Basic Countermeasures}\BBCQ\
\newblock In {\Bem ACM SIGSAC Conference on Computer and Communications
  Security (CCS)}.

\bibitem[\protect\BCAY{Friedrich, Stammer, Schramowski,\ \BBA\
  Kersting}{Friedrich et~al.}{2023}]{RiT}
Friedrich, F., Stammer, W., Schramowski, P., \BBA\ Kersting, K. \BBOP2023\BBCP.
\newblock \BBOQ {Revision Transformers: Instructing Language Models to Change
  their Values}\BBCQ\
\newblock In {\Bem European Conference on Artificial Intelligence (ECAI)}.

\bibitem[\protect\BCAY{{Getty Images}}{{Getty
  Images}}{2023}]{getty_images_lawsuit_statement}
{Getty Images} \BBOP2023\BBCP.
\newblock \BBOQ {Getty Images Statement}\BBCQ\
\newblock GettyImages.
  \url{https://newsroom.gettyimages.com/en/getty-images/getty-images-statement}.
\newblock Online; accessed 27-January-2023.

\bibitem[\protect\BCAY{{Glyn Lowe Photo Works}}{{Glyn Lowe Photo
  Works}}{2014}]{adam_sandler}
{Glyn Lowe Photo Works} \BBOP2014\BBCP.
\newblock \BBOQ {File:Adam Sandler on 'Blended' Red Carpet in Berlin
  (14043442427).jpg}\BBCQ\
\newblock
  \url{https://commons.wikimedia.org/wiki/File:Adam_Sandler_on_\%27Blended\%27_Red_Carpet_in_Berlin_\%2814043442427\%29.jpg}.
\newblock Online; accessed 01-February-2023, licensed under CC BY 2.0, image
  cropped.

\bibitem[\protect\BCAY{Goh, Cammarata, Voss, Carter, Petrov, Schubert,
  Radford,\ \BBA\ Olah}{Goh et~al.}{2021}]{goh2021multimodal}
Goh, G., Cammarata, N., Voss, C., Carter, S., Petrov, M., Schubert, L.,
  Radford, A., \BBA\ Olah, C. \BBOP2021\BBCP.
\newblock \BBOQ {Multimodal Neurons in Artificial Neural Networks}\BBCQ\
\newblock Distill.
\newblock \url{https://distill.pub/2021/multimodal-neurons}.

\bibitem[\protect\BCAY{Goodfellow, Shlens,\ \BBA\ Szegedy}{Goodfellow
  et~al.}{2015}]{adv_attacks_goodfellow}
Goodfellow, I.~J., Shlens, J., \BBA\ Szegedy, C. \BBOP2015\BBCP.
\newblock \BBOQ {Explaining and Harnessing Adversarial Examples}\BBCQ\
\newblock In {\Bem International Conference on Learning Representations,
  {ICLR}}.

\bibitem[\protect\BCAY{He, Zhang, Ren,\ \BBA\ Sun}{He et~al.}{2016}]{he2016}
He, K., Zhang, X., Ren, S., \BBA\ Sun, J. \BBOP2016\BBCP.
\newblock \BBOQ {Deep Residual Learning for Image Recognition}\BBCQ\
\newblock In {\Bem IEEE/CVF Conference on Computer Vision and Pattern
  Recognition (CVPR)}.

\bibitem[\protect\BCAY{Heikkilä}{Heikkilä}{2022}]{what_does_gpt3_know}
Heikkilä, M. \BBOP2022\BBCP.
\newblock \BBOQ {What does GPT-3 “know” about me?}\BBCQ\
\newblock MIT Technology Review.
  \url{https://www.technologyreview.com/2022/08/31/1058800/what-does-gpt-3-know-about-me/}.
\newblock Online; accessed 29-January-2023.

\bibitem[\protect\BCAY{Hernandez}{Hernandez}{2014}]{ilene_kristen}
Hernandez, G. \BBOP2014\BBCP.
\newblock \BBOQ {Ilene Kristen at 2014 Daytime Emmys}\BBCQ\
\newblock \url{https://www.flickr.com/photos/greginhollywood/14353837150/}.
\newblock Online; accessed 01-February-2023, licensed under CC BY 2.0, image
  cropped.

\bibitem[\protect\BCAY{Hintersdorf, Struppek,\ \BBA\ Kersting}{Hintersdorf
  et~al.}{2022}]{hintersdorfMIA}
Hintersdorf, D., Struppek, L., \BBA\ Kersting, K. \BBOP2022\BBCP.
\newblock \BBOQ {To Trust or Not To Trust Prediction Scores for Membership
  Inference Attacks}\BBCQ\
\newblock In {\Bem International Joint Conference on Artificial Intelligence,
  (IJCAI)}.

\bibitem[\protect\BCAY{Hu\ \BBA\ Pang}{Hu\ \BBA\
  Pang}{2023}]{meminf_diffusion_models}
Hu, H.\BBACOMMA\  \BBA\ Pang, J. \BBOP2023\BBCP.
\newblock \BBOQ {Membership Inference of Diffusion Models}\BBCQ\
\newblock {\Bem arXiv preprint}, {\Bem arXiv:2301.09956}.

\bibitem[\protect\BCAY{{IEEE}}{{IEEE}}{2024}]{ieee_style_manual}
{IEEE} \BBOP2024\BBCP.
\newblock \BBOQ {IEEE Editorial Style Manual}\BBCQ\
\newblock
  \url{http://journals.ieeeauthorcenter.ieee.org/wp-content/uploads/sites/7/IEEE-Editorial-Style-Manual-for-Authors.pdf}.

\bibitem[\protect\BCAY{Ilharco, Wortsman, Wightman, Gordon, Carlini, Taori,
  Dave, Shankar, Namkoong, Miller, Hajishirzi, Farhadi,\ \BBA\ Schmidt}{Ilharco
  et~al.}{2021}]{open_clip}
Ilharco, G., Wortsman, M., Wightman, R., Gordon, C., Carlini, N., Taori, R.,
  Dave, A., Shankar, V., Namkoong, H., Miller, J., Hajishirzi, H., Farhadi, A.,
  \BBA\ Schmidt, L. \BBOP2021\BBCP.
\newblock \BBOQ {OpenCLIP}\BBCQ\
\newblock Zenodo. https://doi.org/10.5281/zenodo.5143773.

\bibitem[\protect\BCAY{Jaiswal\ \BBA\ Mower~Provost}{Jaiswal\ \BBA\
  Mower~Provost}{2020}]{multi_modal_emotion}
Jaiswal, M.\BBACOMMA\  \BBA\ Mower~Provost, E. \BBOP2020\BBCP.
\newblock \BBOQ {Privacy Enhanced Multimodal Neural Representations for Emotion
  Recognition}\BBCQ\
\newblock In {\Bem AAAI Conference on Artificial Intelligence}.

\bibitem[\protect\BCAY{JCS}{JCS}{2012a}]{bernhard_hoecker}
JCS \BBOP2012a\BBCP.
\newblock \BBOQ {Bernhard Hoecker beim Deutschen Fernsehpreis 2012}\BBCQ\
\newblock
  \url{https://commons.wikimedia.org/wiki/File:Deutscher_Fernsehpreis_2012_-_Bernhard_Hoecker.jpg}.
\newblock Online; accessed 01-February-2023, licensed under CC BY 3.0, image
  cropped.

\bibitem[\protect\BCAY{JCS}{JCS}{2012b}]{bettina_lamprecht}
JCS \BBOP2012b\BBCP.
\newblock \BBOQ {Bettina Lamprecht und Matthias Matschke beim Deutschen
  Fernsehpreis 2012}\BBCQ\
\newblock
  \url{https://commons.wikimedia.org/wiki/File:Deutscher_Fernsehpreis_2012_-_Bettina_Lamprecht_-_Matthias_Matschke_2.jpg}.
\newblock Online; accessed 01-February-2023, licensed under CC BY 3.0, image
  cropped.

\bibitem[\protect\BCAY{JCS}{JCS}{2015}]{guido_cantz}
JCS \BBOP2015\BBCP.
\newblock \BBOQ {Guido Cantz auf der Frankfurter Buchmesse 2015}\BBCQ\
\newblock
  \url{https://commons.wikimedia.org/wiki/File:Frankfurter_Buchmesse_2015_-_Guido_Cantz_1.JPG}.
\newblock Online; accessed 01-February-2023, licensed under CC BY 3.0, image
  cropped.

\bibitem[\protect\BCAY{JCS}{JCS}{2017}]{max_giermann}
JCS \BBOP2017\BBCP.
\newblock \BBOQ {Max Giermann beim Hessischen Film- und Kinopreis 2017}\BBCQ\
\newblock
  \url{https://commons.wikimedia.org/wiki/File:Hessischer_Filmpreis_2017_-_Max_Giermann_2.JPG}.
\newblock Online; accessed 01-February-2023, licensed under CC BY 3.0, image
  cropped.

\bibitem[\protect\BCAY{Kingma\ \BBA\ Ba}{Kingma\ \BBA\ Ba}{2015}]{adam}
Kingma, D.~P.\BBACOMMA\  \BBA\ Ba, J. \BBOP2015\BBCP.
\newblock \BBOQ {Adam: {A} Method for Stochastic Optimization}\BBCQ\
\newblock In {\Bem International Conference on Learning Representations
  (ICLR)}.

\bibitem[\protect\BCAY{Krichel}{Krichel}{2019}]{carolin_kebekus}
Krichel, H. \BBOP2019\BBCP.
\newblock \BBOQ {Comedian Carolin Kebekus at the SWR3 New Pop Festival
  2019}\BBCQ\
\newblock
  \url{https://commons.wikimedia.org/wiki/File:Carolin_Kebekus-5848.jpg}.
\newblock Online; accessed 01-February-2023, licensed under CC BY 3.0, image
  cropped.

\bibitem[\protect\BCAY{Li, Rezaei,\ \BBA\ Liu}{Li
  et~al.}{2022}]{li2022userlevel}
Li, G., Rezaei, S., \BBA\ Liu, X. \BBOP2022\BBCP.
\newblock \BBOQ {User-Level Membership Inference Attack against Metric
  Embedding Learning}\BBCQ\
\newblock In {\Bem ICLR Workshop on PAIR\textsuperscript{2}Struct: Privacy,
  Accountability, Interpretability, Robustness, Reasoning on Structured Data}.

\bibitem[\protect\BCAY{Li\ \BBA\ Zhang}{Li\ \BBA\ Zhang}{2021}]{li2021}
Li, Z.\BBACOMMA\  \BBA\ Zhang, Y. \BBOP2021\BBCP.
\newblock \BBOQ {Membership Leakage in Label-Only Exposures}\BBCQ\
\newblock In {\Bem ACM SIGSAC Conference on Computer and Communications
  Security (CCS)}.

\bibitem[\protect\BCAY{Liu, Jia, Qu,\ \BBA\ Gong}{Liu et~al.}{2021}]{encodermi}
Liu, H., Jia, J., Qu, W., \BBA\ Gong, N.~Z. \BBOP2021\BBCP.
\newblock \BBOQ {EncoderMI: Membership Inference against Pre-Trained Encoders
  in Contrastive Learning}\BBCQ\
\newblock In {\Bem ACM SIGSAC Conference on Computer and Communications
  Security (CCS)}.

\bibitem[\protect\BCAY{Liu, Luo, Wang,\ \BBA\ Tang}{Liu et~al.}{2015}]{celeba}
Liu, Z., Luo, P., Wang, X., \BBA\ Tang, X. \BBOP2015\BBCP.
\newblock \BBOQ {Deep Learning Face Attributes in the Wild}\BBCQ\
\newblock In {\Bem IEEE International Conference on Computer Vision (ICCV)}.

\bibitem[\protect\BCAY{Lomas}{Lomas}{2021}]{ever_techcrunch}
Lomas, N. \BBOP2021\BBCP.
\newblock \BBOQ {FTC settlement with Ever orders data and AIs deleted after
  facial recognition pivot}\BBCQ\
\newblock TechCrunch.
  \url{https://techcrunch.com/2021/01/12/ftc-settlement-with-ever-orders-data-and-ais-deleted-after-facial-recognition-pivot/}.
\newblock Online; accessed 01-September-2022.

\bibitem[\protect\BCAY{Lomas}{Lomas}{2022a}]{clearview_fine_italy}
Lomas, N. \BBOP2022a\BBCP.
\newblock \BBOQ {Italy fines Clearview AI €20M and orders data deleted}\BBCQ\
\newblock TechCrunch.
  \url{https://techcrunch.com/2022/03/09/clearview-italy-gdpr/}.
\newblock Online; accessed 01-September-2022.

\bibitem[\protect\BCAY{Lomas}{Lomas}{2022b}]{clearview_fine_greece}
Lomas, N. \BBOP2022b\BBCP.
\newblock \BBOQ {Selfie scraping Clearview AI hit with another €20M ban order
  in Europe}\BBCQ\
\newblock TechCrunch.
  \url{https://techcrunch.com/2022/07/13/clearview-greek-ban-order/}.
\newblock Online; accessed 01-September-2022.

\bibitem[\protect\BCAY{Lomas}{Lomas}{2022c}]{clearview_fine_uk}
Lomas, N. \BBOP2022c\BBCP.
\newblock \BBOQ {UK fines Clearview just under \$10M for privacy
  breaches}\BBCQ\
\newblock TechCrunch.
  \url{https://techcrunch.com/2022/05/23/clearview-uk-ico-fine/}.
\newblock Online; accessed 01-September-2022.

\bibitem[\protect\BCAY{Maggie}{Maggie}{2007}]{valerie_harper}
Maggie \BBOP2007\BBCP.
\newblock \BBOQ {Valerie}\BBCQ\
\newblock \url{https://www.flickr.com/photos/38494596@N00/350140882/}.
\newblock Online; accessed 01-February-2023, licensed under CC BY 2.0, image
  cropped.

\bibitem[\protect\BCAY{Miao, Xue, Chen, Pan, Zhang, Zhao, Kaafar,\ \BBA\
  Xiang}{Miao et~al.}{2021}]{maio_user_level_audio}
Miao, Y., Xue, M., Chen, C., Pan, L., Zhang, J., Zhao, B. Z.~H., Kaafar, D.,
  \BBA\ Xiang, Y. \BBOP2021\BBCP.
\newblock \BBOQ {The Audio Auditor: User-Level Membership Inference in Internet
  of Things Voice Services}\BBCQ\
\newblock In {\Bem Proceedings on Privacy Enhancing Technologies Symposium
  (PoPETs)}.

\bibitem[\protect\BCAY{Milli{\`{e}}re}{Milli{\`{e}}re}{2022}]{milliere2022}
Milli{\`{e}}re, R. \BBOP2022\BBCP.
\newblock \BBOQ {Adversarial Attacks on Image Generation With Made-Up
  Words}\BBCQ\
\newblock {\Bem arXiv preprint}, {\Bem arXiv:2208.04135}.

\bibitem[\protect\BCAY{Mokady, Hertz,\ \BBA\ Bermano}{Mokady
  et~al.}{2021}]{clip_captioning}
Mokady, R., Hertz, A., \BBA\ Bermano, A.~H. \BBOP2021\BBCP.
\newblock \BBOQ {{ClipCap}: {CLIP} Prefix for Image Captioning}\BBCQ\
\newblock {\Bem arXiv preprint}, {\Bem arXiv:2111.09734}.

\bibitem[\protect\BCAY{{Montclair Film}}{{Montclair Film}}{2013}]{jimmy_fallon}
{Montclair Film} \BBOP2013\BBCP.
\newblock \BBOQ {TTL\_4528}\BBCQ\
\newblock \url{https://www.flickr.com/photos/montclairfilmfest/11046018105/}.
\newblock Online; accessed 28-January-2023, licensed under CC BY 2.0, image
  cropped.

\bibitem[\protect\BCAY{{Montclair Film (Photography by Neil
  Grabowsky)}}{{Montclair Film (Photography by Neil
  Grabowsky)}}{2019}]{ben_stiller}
{Montclair Film (Photography by Neil Grabowsky)} \BBOP2019\BBCP.
\newblock \BBOQ {In Conversation with Ben Stiller}\BBCQ\
\newblock \url{https://www.flickr.com/photos/montclairfilmfest/46998980874/}.
\newblock Online; accessed 28-January-2023, licensed under CC BY 2.0, image
  cropped.

\bibitem[\protect\BCAY{Ng\ \BBA\ Winkler}{Ng\ \BBA\ Winkler}{2014}]{facescrub}
Ng, H.-W.\BBACOMMA\  \BBA\ Winkler, S. \BBOP2014\BBCP.
\newblock \BBOQ {A data-driven approach to cleaning large face datasets}\BBCQ\
\newblock In {\Bem IEEE International Conference on Image Processing (ICIP)}.

\bibitem[\protect\BCAY{Nichol, Dhariwal, Ramesh, Shyam, Mishkin, Mcgrew,
  Sutskever,\ \BBA\ Chen}{Nichol et~al.}{2022}]{glide}
Nichol, A.~Q., Dhariwal, P., Ramesh, A., Shyam, P., Mishkin, P., Mcgrew, B.,
  Sutskever, I., \BBA\ Chen, M. \BBOP2022\BBCP.
\newblock \BBOQ {{GLIDE}: Towards Photorealistic Image Generation and Editing
  with Text-Guided Diffusion Models}\BBCQ\
\newblock In {\Bem International Conference on Machine Learning (ICML)}.

\bibitem[\protect\BCAY{OpenAI}{OpenAI}{2023}]{gpt4}
OpenAI \BBOP2023\BBCP.
\newblock \BBOQ {{GPT-4} Technical Report}\BBCQ\
\newblock {\Bem arXiv preprint}, {\Bem arXiv:2301.09956}.

\bibitem[\protect\BCAY{Ouyang, Wu, Jiang, Almeida, Wainwright, Mishkin, Zhang,
  Agarwal, Slama, Ray, Schulman, Hilton, Kelton, Miller, Simens, Askell,
  Welinder, Christiano, Leike,\ \BBA\ Lowe}{Ouyang et~al.}{2022}]{instruct_gpt}
Ouyang, L., Wu, J., Jiang, X., Almeida, D., Wainwright, C., Mishkin, P., Zhang,
  C., Agarwal, S., Slama, K., Ray, A., Schulman, J., Hilton, J., Kelton, F.,
  Miller, L., Simens, M., Askell, A., Welinder, P., Christiano, P.~F., Leike,
  J., \BBA\ Lowe, R. \BBOP2022\BBCP.
\newblock \BBOQ {Training language models to follow instructions with human
  feedback}\BBCQ\
\newblock In {\Bem Advances in Neural Information Processing Systems
  (NeurIPS)}.

\bibitem[\protect\BCAY{{Parliament of Canada}}{{Parliament of
  Canada}}{2000}]{pipeda_canada}
{Parliament of Canada} \BBOP2000\BBCP.
\newblock \BBOQ {Personal Information Protection and Electronic Documents
  Act}\BBCQ\
\newblock \url{https://laws-lois.justice.gc.ca/ENG/ACTS/P-8.6/FullText.html}.
\newblock Online; accessed 22-January-2023.

\bibitem[\protect\BCAY{Radford, Kim, Hallacy, Ramesh, Goh, Agarwal, Sastry,
  Askell, Mishkin, Clark, Krueger,\ \BBA\ Sutskever}{Radford
  et~al.}{2021}]{clip_radford}
Radford, A., Kim, J.~W., Hallacy, C., Ramesh, A., Goh, G., Agarwal, S., Sastry,
  G., Askell, A., Mishkin, P., Clark, J., Krueger, G., \BBA\ Sutskever, I.
  \BBOP2021\BBCP.
\newblock \BBOQ {Learning Transferable Visual Models From Natural Language
  Supervision}\BBCQ\
\newblock In {\Bem International Conference on Machine Learning (ICML)}.

\bibitem[\protect\BCAY{Rahman, Fritz, Backes,\ \BBA\ Zhang}{Rahman
  et~al.}{2020}]{everything_about_you}
Rahman, T.~A., Fritz, M., Backes, M., \BBA\ Zhang, Y. \BBOP2020\BBCP.
\newblock \BBOQ {Everything About You: {A} Multimodal Approach towards
  Friendship Inference in Online Social Networks}\BBCQ\
\newblock {\Bem arXiv preprint}, {\Bem arXiv:2003.00996}.

\bibitem[\protect\BCAY{Ramesh, Dhariwal, Nichol, Chu,\ \BBA\ Chen}{Ramesh
  et~al.}{2022}]{dalle2}
Ramesh, A., Dhariwal, P., Nichol, A., Chu, C., \BBA\ Chen, M. \BBOP2022\BBCP.
\newblock \BBOQ {Hierarchical Text-Conditional Image Generation with {CLIP}
  Latents}\BBCQ\
\newblock {\Bem arXiv preprint}, {\Bem arXiv:2204.06125}.

\bibitem[\protect\BCAY{Ramesh, Pavlov, Goh, Gray, Voss, Radford, Chen,\ \BBA\
  Sutskever}{Ramesh et~al.}{2021}]{dalle}
Ramesh, A., Pavlov, M., Goh, G., Gray, S., Voss, C., Radford, A., Chen, M.,
  \BBA\ Sutskever, I. \BBOP2021\BBCP.
\newblock \BBOQ {Zero-Shot Text-to-Image Generation}\BBCQ\
\newblock In {\Bem International Conference on Machine Learning (ICML)}.

\bibitem[\protect\BCAY{Rezaei\ \BBA\ Liu}{Rezaei\ \BBA\
  Liu}{2021}]{rezaei2021difficulty}
Rezaei, S.\BBACOMMA\  \BBA\ Liu, X. \BBOP2021\BBCP.
\newblock \BBOQ {On the Difficulty of Membership Inference Attacks}\BBCQ\
\newblock In {\Bem IEEE/CVF Conference on Computer Vision and Pattern
  Recognition (CVPR)}.

\bibitem[\protect\BCAY{Rombach, Blattmann, Lorenz, Esser,\ \BBA\ Ommer}{Rombach
  et~al.}{2022}]{rombach_diffusion}
Rombach, R., Blattmann, A., Lorenz, D., Esser, P., \BBA\ Ommer, B.
  \BBOP2022\BBCP.
\newblock \BBOQ {High-Resolution Image Synthesis With Latent Diffusion
  Models}\BBCQ\
\newblock In {\Bem IEEE/CVF Conference on Computer Vision and Pattern
  Recognition (CVPR)}.

\bibitem[\protect\BCAY{Russakovsky, Deng, Su, Krause, Satheesh, Ma, Huang,
  Karpathy, Khosla, Bernstein, Berg,\ \BBA\ Fei-Fei}{Russakovsky
  et~al.}{2015}]{imagenet}
Russakovsky, O., Deng, J., Su, H., Krause, J., Satheesh, S., Ma, S., Huang, Z.,
  Karpathy, A., Khosla, A., Bernstein, M., Berg, A.~C., \BBA\ Fei-Fei, L.
  \BBOP2015\BBCP.
\newblock \BBOQ {ImageNet Large Scale Visual Recognition Challenge}\BBCQ\
\newblock {\Bem International Journal of Computer Vision (IJCV)}, {\Bem 115},
  211--252.

\bibitem[\protect\BCAY{Saharia, Chan, Saxena, Li, Whang, Denton, Ghasemipour,
  Gontijo-Lopes, Ayan, Salimans, Ho, Fleet,\ \BBA\ Norouzi}{Saharia
  et~al.}{2022}]{imagen}
Saharia, C., Chan, W., Saxena, S., Li, L., Whang, J., Denton, E., Ghasemipour,
  S. K.~S., Gontijo-Lopes, R., Ayan, B.~K., Salimans, T., Ho, J., Fleet, D.~J.,
  \BBA\ Norouzi, M. \BBOP2022\BBCP.
\newblock \BBOQ {Photorealistic Text-to-Image Diffusion Models with Deep
  Language Understanding}\BBCQ\
\newblock In {\Bem Advances in Neural Information Processing Systems
  (NeurIPS)}.

\bibitem[\protect\BCAY{Salem, Zhang, Humbert, Berrang, Fritz,\ \BBA\
  Backes}{Salem et~al.}{2019}]{salem2019}
Salem, A., Zhang, Y., Humbert, M., Berrang, P., Fritz, M., \BBA\ Backes, M.
  \BBOP2019\BBCP.
\newblock \BBOQ {ML-Leaks: Model and Data Independent Membership Inference
  Attacks and Defenses on Machine Learning Models}\BBCQ\
\newblock In {\Bem Network and Distributed System Security Symposium (NDSS)}.

\bibitem[\protect\BCAY{Sandoval}{Sandoval}{2009}]{kaley_cuoco}
Sandoval, M.~J. \BBOP2009\BBCP.
\newblock \BBOQ {Jim Parsons, Kaley Cuoco (The Big Bang Theory)}\BBCQ\
\newblock \url{https://www.flickr.com/photos/therainstopped/3781595029/}.
\newblock Online; accessed 28-January-2023, licensed under CC BY 2.0, image
  cropped.

\bibitem[\protect\BCAY{Schick\ \BBA\ Sch{\"u}tze}{Schick\ \BBA\
  Sch{\"u}tze}{2021}]{iPET}
Schick, T.\BBACOMMA\  \BBA\ Sch{\"u}tze, H. \BBOP2021\BBCP.
\newblock \BBOQ {It{'}s Not Just Size That Matters: Small Language Models Are
  Also Few-Shot Learners}\BBCQ\
\newblock In {\Bem Conference of the North American Chapter of the Association
  for Computational Linguistics: Human Language Technologies (NAACL)}.

\bibitem[\protect\BCAY{Schramowski, Tauchmann,\ \BBA\ Kersting}{Schramowski
  et~al.}{2022}]{schramowski2022}
Schramowski, P., Tauchmann, C., \BBA\ Kersting, K. \BBOP2022\BBCP.
\newblock \BBOQ {Can Machines Help Us Answering Question 16 in Datasheets, and
  In Turn Reflecting on Inappropriate Content?}\BBCQ\
\newblock In {\Bem ACM Conference on Fairness, Accountability, and Transparency
  (FAccT)}.

\bibitem[\protect\BCAY{Schuhmann, Beaumont, Vencu, Gordon, Wightman, Cherti,
  Coombes, Katta, Mullis, Wortsman, Schramowski, Kundurthy, Crowson, Schmidt,
  Kaczmarczyk,\ \BBA\ Jitsev}{Schuhmann et~al.}{2022}]{laion5b}
Schuhmann, C., Beaumont, R., Vencu, R., Gordon, C., Wightman, R., Cherti, M.,
  Coombes, T., Katta, A., Mullis, C., Wortsman, M., Schramowski, P., Kundurthy,
  S., Crowson, K., Schmidt, L., Kaczmarczyk, R., \BBA\ Jitsev, J.
  \BBOP2022\BBCP.
\newblock \BBOQ {LAION-5B: An open large-scale dataset for training next
  generation image-text models}\BBCQ\
\newblock In {\Bem Advances in Neural Information Processing Systems
  (NeurIPS)}.

\bibitem[\protect\BCAY{Schuhmann, Vencu, Beaumont, Kaczmarczyk, Mullis, Katta,
  Coombes, Jitsev,\ \BBA\ Komatsuzaki}{Schuhmann et~al.}{2021}]{laion400m}
Schuhmann, C., Vencu, R., Beaumont, R., Kaczmarczyk, R., Mullis, C., Katta, A.,
  Coombes, T., Jitsev, J., \BBA\ Komatsuzaki, A. \BBOP2021\BBCP.
\newblock \BBOQ {{LAION-400M:} Open Dataset of CLIP-Filtered 400 Million
  Image-Text Pairs}\BBCQ\
\newblock In {\Bem NeurIPS Data-Centric AI Workshop}.

\bibitem[\protect\BCAY{Sharma, Ding, Goodman,\ \BBA\ Soricut}{Sharma
  et~al.}{2018}]{cc3m}
Sharma, P., Ding, N., Goodman, S., \BBA\ Soricut, R. \BBOP2018\BBCP.
\newblock \BBOQ {Conceptual Captions: A Cleaned, Hypernymed, Image Alt-text
  Dataset For Automatic Image Captioning}\BBCQ\
\newblock In {\Bem Annual Meeting of the Association for Computational
  Linguistics (ACL)}.

\bibitem[\protect\BCAY{Shokri, Stronati, Song,\ \BBA\ Shmatikov}{Shokri
  et~al.}{2017}]{shokri2017}
Shokri, R., Stronati, M., Song, C., \BBA\ Shmatikov, V. \BBOP2017\BBCP.
\newblock \BBOQ {Membership Inference Attacks Against Machine Learning
  Models}\BBCQ\
\newblock In {\Bem IEEE Symposium on Security and Privacy (S\&P)}.

\bibitem[\protect\BCAY{Song\ \BBA\ Shmatikov}{Song\ \BBA\
  Shmatikov}{2019}]{song_user_level_text}
Song, C.\BBACOMMA\  \BBA\ Shmatikov, V. \BBOP2019\BBCP.
\newblock \BBOQ {Auditing Data Provenance in Text-Generation Models}\BBCQ\
\newblock In {\Bem ACM SIGKDD International Conference on Knowledge Discovery
  \& Data Mining}.

\bibitem[\protect\BCAY{Struppek, Hintersdorf, De~Almeida~Correira, Adler,\
  \BBA\ Kersting}{Struppek et~al.}{2022a}]{struppekModelInv}
Struppek, L., Hintersdorf, D., De~Almeida~Correira, A., Adler, A., \BBA\
  Kersting, K. \BBOP2022a\BBCP.
\newblock \BBOQ {Plug \& Play Attacks: Towards Robust and Flexible Model
  Inversion Attacks}\BBCQ\
\newblock In {\Bem International Conference on Machine Learning (ICML)}.

\bibitem[\protect\BCAY{Struppek, Hintersdorf, Neider,\ \BBA\ Kersting}{Struppek
  et~al.}{2022b}]{struppek_neuralhash}
Struppek, L., Hintersdorf, D., Neider, D., \BBA\ Kersting, K. \BBOP2022b\BBCP.
\newblock \BBOQ {Learning to Break Deep Perceptual Hashing: The Use Case
  NeuralHash}\BBCQ\
\newblock In {\Bem ACM Conference on Fairness, Accountability, and Transparency
  (FAccT)}.

\bibitem[\protect\BCAY{Szegedy, Zaremba, Sutskever, Bruna, Erhan, Goodfellow,\
  \BBA\ Fergus}{Szegedy et~al.}{2014}]{szegedy_2014}
Szegedy, C., Zaremba, W., Sutskever, I., Bruna, J., Erhan, D., Goodfellow,
  I.~J., \BBA\ Fergus, R. \BBOP2014\BBCP.
\newblock \BBOQ {Intriguing properties of neural networks}\BBCQ\
\newblock In {\Bem International Conference on Learning Representations
  (ICLR)}.

\bibitem[\protect\BCAY{Tan\ \BBA\ Le}{Tan\ \BBA\ Le}{2019}]{efficient_net}
Tan, M.\BBACOMMA\  \BBA\ Le, Q. \BBOP2019\BBCP.
\newblock \BBOQ {{E}fficient{N}et: Rethinking Model Scaling for Convolutional
  Neural Networks}\BBCQ\
\newblock In {\Bem International Conference on Machine Learning (ICML)}.

\bibitem[\protect\BCAY{Thies}{Thies}{2022}]{clip_me_if_you_can}
Thies, J. \BBOP2022\BBCP.
\newblock \BBOQ {CLIPme if you can!}\BBCQ\
\newblock \url{https://justusthies.github.io/posts/clipme/}.
\newblock Online; accessed 10-August-2022.

\bibitem[\protect\BCAY{Trilling\ \BBA\ Wetherill}{Trilling\ \BBA\
  Wetherill}{2021}]{ever_ftc}
Trilling, J.\BBACOMMA\  \BBA\ Wetherill, R. \BBOP2021\BBCP.
\newblock \BBOQ {California Company Settles FTC Allegations It Deceived
  Consumers about use of Facial Recognition in Photo Storage App}\BBCQ\
\newblock FTC.
  \url{https://www.ftc.gov/news-events/news/press-releases/2021/01/california-company-settles-ftc-allegations-it-deceived-consumers-about-use-facial-recognition-photo}.
\newblock Online; accessed 01-September-2022.

\bibitem[\protect\BCAY{{United States Census Bureau}}{{United States Census
  Bureau}}{2021}]{last_names_list}
{United States Census Bureau} \BBOP2021\BBCP.
\newblock \BBOQ {Frequently Occurring Surnames from the 2010 Census}\BBCQ\
\newblock
  \url{https://www.census.gov/topics/population/genealogy/data/2010_surnames.html}.
\newblock Online; accessed 06-February-2023.

\bibitem[\protect\BCAY{Vaswani, Shazeer, Parmar, Uszkoreit, Jones, Gomez,
  Kaiser,\ \BBA\ Polosukhin}{Vaswani et~al.}{2017}]{attention_is_all_you_need}
Vaswani, A., Shazeer, N., Parmar, N., Uszkoreit, J., Jones, L., Gomez, A.~N.,
  Kaiser, L., \BBA\ Polosukhin, I. \BBOP2017\BBCP.
\newblock \BBOQ {Attention is All you Need}\BBCQ\
\newblock In {\Bem Advances in Neural Information Processing Systems
  (NeurIPS)}.

\bibitem[\protect\BCAY{Vincent}{Vincent}{2023}]{getty_images_lawsuit_verge}
Vincent, J. \BBOP2023\BBCP.
\newblock \BBOQ {Getty Images is suing the creators of AI art tool Stable
  Diffusion for scraping its content}\BBCQ\
\newblock The Verge.
  \url{https://www.theverge.com/2023/1/17/23558516/ai-art-copyright-stable-diffusion-getty-images-lawsuit}.
\newblock Online; accessed 27-January-2023.

\bibitem[\protect\BCAY{Wang, FU, Li, Khisti, Zemel,\ \BBA\ Makhzani}{Wang
  et~al.}{2021}]{wang2021}
Wang, K.-C., FU, Y., Li, K., Khisti, A., Zemel, R., \BBA\ Makhzani, A.
  \BBOP2021\BBCP.
\newblock \BBOQ {Variational Model Inversion Attacks}\BBCQ\
\newblock In {\Bem Advances in Neural Information Processing Systems
  (NeurIPS)}.

\bibitem[\protect\BCAY{Webster, Rabin, Simon,\ \BBA\ Jurie}{Webster
  et~al.}{2021}]{this_person_exists}
Webster, R., Rabin, J., Simon, L., \BBA\ Jurie, F. \BBOP2021\BBCP.
\newblock \BBOQ {This Person (Probably) Exists. Identity Membership Attacks
  Against {GAN} Generated Faces}\BBCQ\
\newblock {\Bem arXiv preprint}, {\Bem arXiv:2107.06018}.

\bibitem[\protect\BCAY{Wickham}{Wickham}{2009}]{first_names_list}
Wickham, H. \BBOP2009\BBCP.
\newblock \BBOQ {data-baby-names}\BBCQ\
\newblock GitHub. \url{https://github.com/hadley/data-baby-names}.
\newblock Online; accessed 06-February-2023.

\bibitem[\protect\BCAY{Wu, Yu, Li, Backes,\ \BBA\ Zhang}{Wu
  et~al.}{2022}]{meminf_on_tti}
Wu, Y., Yu, N., Li, Z., Backes, M., \BBA\ Zhang, Y. \BBOP2022\BBCP.
\newblock \BBOQ {Membership Inference Attacks Against Text-to-image Generation
  Models}\BBCQ\
\newblock {\Bem arXiv preprint}, {\Bem arXiv:2210.00968}.

\bibitem[\protect\BCAY{Yeom, Giacomelli, Fredrikson,\ \BBA\ Jha}{Yeom
  et~al.}{2018}]{yeom2018}
Yeom, S., Giacomelli, I., Fredrikson, M., \BBA\ Jha, S. \BBOP2018\BBCP.
\newblock \BBOQ {Privacy risk in machine learning: Analyzing the connection to
  overfitting}\BBCQ\
\newblock In {\Bem IEEE Computer Security Foundations Symposium (CSF)}.

\bibitem[\protect\BCAY{Yu, Xu, Koh, Luong, Baid, Wang, Vasudevan, Ku, Yang,
  Ayan, Hutchinson, Han, Parekh, Li, Zhang, Baldridge,\ \BBA\ Wu}{Yu
  et~al.}{2022}]{parti}
Yu, J., Xu, Y., Koh, J.~Y., Luong, T., Baid, G., Wang, Z., Vasudevan, V., Ku,
  A., Yang, Y., Ayan, B.~K., Hutchinson, B., Han, W., Parekh, Z., Li, X.,
  Zhang, H., Baldridge, J., \BBA\ Wu, Y. \BBOP2022\BBCP.
\newblock \BBOQ {Scaling Autoregressive Models for Content-Rich Text-to-Image
  Generation}\BBCQ\
\newblock {\Bem Transactions on Machine Learning Research}, {\Bem 2022}, 1--53.

\bibitem[\protect\BCAY{Zhang, Li, Huang,\ \BBA\ Yin}{Zhang
  et~al.}{2021}]{retrieval_pip}
Zhang, P.-F., Li, Y., Huang, Z., \BBA\ Yin, H. \BBOP2021\BBCP.
\newblock \BBOQ {Privacy Protection in Deep Multi-Modal Retrieval}\BBCQ\
\newblock In {\Bem ACM SIGIR Conference on Research and Development in
  Information Retrieval}.

\bibitem[\protect\BCAY{Zhang, Jia, Pei, Wang, Li,\ \BBA\ Song}{Zhang
  et~al.}{2020}]{zhang2020}
Zhang, Y., Jia, R., Pei, H., Wang, W., Li, B., \BBA\ Song, D. \BBOP2020\BBCP.
\newblock \BBOQ {The Secret Revealer: Generative Model-Inversion Attacks
  Against Deep Neural Networks}\BBCQ\
\newblock In {\Bem IEEE/CVF Conference on Computer Vision and Pattern
  Recognition (CVPR)}.

\end{thebibliography}
\bibliographystyle{theapa.bst}

\end{document}